\documentclass[journal]{IEEEtran}
\usepackage{graphicx}
\usepackage{amssymb}
\usepackage{amsmath}
\usepackage{epstopdf}
\usepackage{subfig}
\usepackage{stfloats}
\usepackage{microtype}
\usepackage{units}
\usepackage{algorithm}
\usepackage[noend]{algpseudocode}
\usepackage{url}
\usepackage{bm}
\usepackage{multirow}

\usepackage{float}
\usepackage{adjustbox}
\usepackage{tabularx}

\makeatletter
\newcommand{\multiline}[1]{%
  \begin{tabularx}{\dimexpr\linewidth-\ALG@thistlm}[t]{@{}X@{}}
    #1
  \end{tabularx}
}
\makeatother
\usepackage{verbatim}

\graphicspath{{figures/}}
\begin{document}

\title{Is rotation forest the best classifier for problems with continuous features?}
\author{A. Bagnall and M. Flynn and J. Large and J. Lines and A. Bostrom and G. Cawley
\thanks{School of Computing Sciences, University of East Anglia, Norwich, NR4 7TJ, UK.}}
\maketitle
\begin{abstract}
In short, our experiments suggest that yes, on average, rotation forest is better than the most common alternatives when all the attributes are real-valued. Rotation forest is a tree based ensemble that performs transforms on subsets of attributes prior to constructing each tree. We present an empirical comparison of classifiers for problems with only real-valued features. We evaluate classifiers from three families of algorithms: support vector machines; tree-based ensembles; and neural networks tuned with a large grid search.
We compare classifiers on unseen data based on the quality of the decision rule (using classification error) the ability to rank cases (area under the receiver operating characteristic) and the probability estimates (using negative log likelihood). We conclude that, in answer to the question posed in the title, yes, rotation forest is significantly more accurate on average than competing techniques when compared on three distinct sets of datasets.  Further, we assess the impact of the design features of rotation forest through an ablative study that transforms random forest into rotation forest. We identify the major limitation of rotation forest as its scalability, particularly in number of attributes. To overcome this problem we develop a model to predict the train time of the algorithm and hence propose a contract version of rotation forest where a run time cap is imposed {\em a priori}. We demonstrate that on large problems rotation forest can be made an order of magnitude faster without significant loss of accuracy. We also show that there is no real benefit (on average) from tuning rotation forest. We maintain that without any domain knowledge to indicate an algorithm preference, rotation forest should be the default algorithm of choice for problems with continuous attributes.

\end{abstract}

\section{Introduction}

Classification is an intrinsically practical exercise, and our interest is in answering the following question: if we have a new classification problem or set of problems, what family of models should we use given our computational constraints? This interest has arisen from our work in the domain of time series classification~\cite{bagnall17bakeoff}, and through working with many industrial partners, but we cannot find an acceptable answer in the literature. The comparative studies of classifiers give some indication (for example~\cite{delgado14hundreds,kazakov19mlaut}), but most people make the decision for pragmatic or dogmatic reasons. Broadly speaking, there are three families of algorithm that could claim to be state of the art: support vector machines; multilayer perceptrons/deep learning; and tree-based ensembles. Our experience has shown that one algorithm, the tree-based ensemble rotation forest~\cite{rodriguez06rotf}, consistently outperforms other classifiers on data where the attributes are real-valued. Our primary contribution is the test of the hypothesis whether on average, for problems with real-valued attributes, rotation forest is significantly more accurate than other classification algorithms. The evidence of our experimentation supports our core hypothesis: on average, rotation forest outperforms the best of the competing algorithms.

Comparative studies such as this are hard to perform, not least because it is easy to find grounds for criticism. Our choice of algorithms to compare against was guided by the conclusions made in other comparative studies~\cite{delgado14hundreds,kazakov19mlaut,wainer16binary}. We compare rotation forest to: random forest; support vector machines with quadratic and radial basis kernels; neural networks with one and two hidden layers; and gradient boosting.

Other grounds for criticism of comparisons is the performance statistic and the data used to measure performance. We compare based on error, balanced error, negative log likelihood and area under the ROC curve on three sets of datasets containing approximately 200 classification problems. All of these problems have no missing values and only real-valued attributes.

Finally, another tricky issue in comparing classifiers is the problem of model selection and tuning of parameters. We adopt the same methodology for tuning all classifiers. We grid search approximately 1000 parameter settings for each classifier and use a ten fold cross-validation on the train data to assess each parameter combination. We only ever evaluate on the test dataset once with a single model built on the whole train data with the parameter values found to have the lowest cross-validation error on the train data \cite{Cawley2010b}.

We stress that we are not suggesting that rotation forest is the best classifier for all problems of this type. Instead, we maintain that it is better on average. Hence, if no other domain knowledge is available, and it is computationally feasible to build a rotation forest, we believe it should be the starting point for trying to solve any new classification problem with real-valued attributes.

Although the original rotation forest paper has received over 1000 citations (according to Google Scholar), it has had nothing like the attention of other classification algorithms. For example, Breiman's original random forest paper~\cite{breiman01randomforest} has over 35,000 citations and a paper proposing a method of choosing parameters for support vector machines~\cite{chapelle02choosing} has received nearly 3000 citations. If our core hypothesis is correct, why then do not more people use rotation forest and why has there been so little development of the algorithm? We believe there are three reasons. Firstly, rotation forest is not available in machine learning toolkits such as scikit-learn, hence the recent boom in machine learning has passed it by. We provide a basic scikit implementation to help overcome this problem\footnote{www.timeseriesclassification.com/rotationForest.php}. Secondly, the original description of the algorithm set the default number of trees to ten, and this is the default value used in the Weka implementation. It was used with the default values in a recent comparison~\cite{delgado14hundreds} and did not do particularly well. Rotation forest performs significantly better when used with a larger number of trees. Thirdly, the design of rotation forest makes it scale poorly, particularly for problems with a large number of attributes. This is caused by the fact that it always uses all attributes for every tree in the ensemble.

The rest of the paper is structured as follows. Section~\ref{section:background} provides a brief overview of the considered classification algorithm types and Section~\ref{section:rotf} gives a more in-depth description of rotation forest. We describe our experimental design and the datasets used in Section~\ref{section:experiments}. Section~\ref{section:comparison} describes the comparison of rotation forest to a range of alternative algorithms on three sets of datasets and presents the evidence to support our core hypothesis that rotation forest is better on problems with real valued attributes. The remainder of the paper is then concerned with exploring why rotation forest performs better, what its weaknesses are and how can we address them. Section~\ref{section:ablative} assesses the influence of structural differences between random forest and rotation forest in an ablative study.  We then examine rotation forest's sensitivity to parameter values in Section~\ref{section:sensitivity} before assessing the time complexity and proposing an alternative version of rotation forest that attempts to construct the best model within a time constraint in Section~\ref{section:timings}. We conclude in Section~\ref{section:conclusions}.

\section{Background}
\label{section:background}
\subsection{Tree-Based Ensembles}

Tree-based homogeneous ensembles are popular classifiers due to their simplicity and general effectiveness. Popular homogeneous ensemble algorithms based on sampling cases or attributes include: bagging decision trees~\cite{breiman96bagging}; random committee, a technique that creates diversity through randomising the base classifiers, which are a form of random tree; dagging~\cite{ting97dagging}; random forest~\cite{breiman01randomforest}, which combines bootstrap sampling with random attribute selection to construct a collection of unpruned trees; and  rotation forest~\cite{rodriguez06rotf}. Of these, we think it fair to say random forest is by far the most popular, and previous studies have claimed it to be amongst the most accurate of all classifiers~\cite{delgado14hundreds}. By default, in most implementations, these methods combine outputs through a majority vote scheme, which assigns an equal weight to the output of each model. We describe both random forest and rotation forest in more detail below.

Boosting ensemble algorithms seek diversity through iteratively re-weighting the training cases and are also very popular.  These include AdaBoost (adaptive boosting)~\cite{freund96experiments}, which iteratively re-weights based on the training error of the base classifier; multiboost~\cite{webb00multiboosting}, a combination of a boosting strategy (similar to AdaBoost) and wagging, a Poisson weighted form of  Bagging; LogitBoost~\cite{friedman98logitboost} which employs a form of additive logistic regression; and gradient boosting algorithms~\cite{friedman01greedy}, which have become popular through the performance of recent incarnations such as XGBoost~\cite{chen16xgboost}. Boosting algorithms also produce a weighting for each classifier. This weight is usually derived from the training process of the base classifier, which may involve regularisation if cross-validation is not used.

\subsection{Neural networks}

Multi-Layer Perceptron (MLP) neural networks are well established as a modern
statistical pattern recognition method (for an overview, see~\cite{Bishop1995a}).  Unlike
a single layer perceptron network that is only capable of solving linearly separable
problems, multi-layer networks, with two or three layers of modifiable connections
are able to solve pattern recognition tasks with convex, concave or disjoint decision
regions.  In order to limit the complexity of the model, and hence avoid over-fitting,
it is common to use some form of regularisation, often the simple weight-decay
regulariser.  The usual regularisation parameter can be tuned by minimising the cross-validation error, or via a Bayesian approach that maximises the
marginal likelihood (also known as the Bayesian evidence for the model), following the
approach of MacKay~\cite{MacKay1992d}. Unlike a single layer perceptron network, where the optimal weights can be determined in closed form, MLPs are typically trained using gradient descent methods, with gradient information calculated using the back-propagation algorithm.  For the small to medium-sized networks considered here, scaled conjugate gradient descent is generally effective.

\subsection{Support Vector Machines}

Kernel learning methods \cite{Scholkopf2002a}, and the support vector machine (SVM)
\cite{Boser1992a,Cortes1995a} in particular have attracted considerable interest due to
a combination of state-of-the-art performance on a variety of real world tasks, and
mathematical tractability, which facilitated considerable theoretical support from
computational learning theory \cite{Vapnik1998a}.  The SVM constructs a linear
maximum-margin classifier \cite{Boser1992a} in a feature space given by a non-linear
transformation of the attributes.  However, rather than specify this transformation
directly, it is implicitly given by a kernel function, that gives the inner products
between vectors in the feature space.  A variety of kernel functions have been suggested,
but the most common are the linear, polynomial and radial basis function (RBF) kernels.
The RBF kernel is a common choice as this gives a classifier capable of arbitrarily
complex decision regions, and again avoidance of over-fitting comes down to careful
tuning of the regularisation and kernel parameters (often achieved via minimisation of
a cross-validation or performance bound-based model selection criterion \cite{Hsu2010a}).
One advantage the support vector machine holds over the multi-layer neural network is
that the training criterion is convex, and hence has a single, global optimum.  Efficient
training algorithms, such as sequential minimal optimisation \cite{Platt1999a}, have been
developed and implemented in freely available software packages, such as LibSVM
\cite{CC01a}.

\section{The Rotation forest algorithm}
\label{section:rotf}

Rotation forest is a tree-based ensemble with some key differences to random forest. The two main differences are that rotation forest transforms the attributes into sets of principle components, and that it uses a C4.5 decision tree. There are more subtle differences, and we describe these through deconstructing the algorithm, shown in pseudocode in Algorithm~\ref{algo:RotF}. For each tree, the transformation process is more complex than might be imagined. Firstly, it uses all attributes for each tree, in contrast to the random forest approach of sampling attributes at each node of the tree. Attributes are split into $r$ random sets of a given size $f$ (step 3), and the transformation is built independently for each set of attributes. However, there is a further step before the transformation which involves discarding instances. Firstly, instances of a given class may be discarded (step 6). Secondly, each group is then sampled with replacement to include a given proportion of cases (step 7). A PCA model is then built on this reduced dataset, and the model then applied to all instances to generate $f$ new attributes for that particular set (step 9). The new attributes are then assembled to form a new dataset with $m=r \cdot f$ attributes.
\begin{algorithm}[!ht]
	\caption{buildRotationForest(Data $D$)}
\label{algo:RotF}
	\begin{algorithmic}[1]
\Require $k$, the number of trees, $f$, the number of features, $p$, the sample proportion
\State Let ${\bf F}=<F_1 \ldots F_k>$ be the C4.5 trees in the forest.
\For{$i \leftarrow  1$ to $k$}
\State \multiline{Randomly partition the original features into $r$ subsets, each with $f$ features ($r=m/f$), denoted $<S_1 \ldots S_r>$.}
\State Let $D_i$ be the train set for tree $i$, initialised to the original data, $D_i \leftarrow D$.
\For {$j \leftarrow 1$ to $r$}
\State \multiline{Select a non-empty subset of classes and extract only cases with those class labels. Each class has 0.5 probability of inclusion.}
\State \multiline{Draw a proportion $p$ of cases (without replacement) of those with the selected class value}
\State \multiline{Perform a Principal Component Analysis (PCA) on the features in $S_j$ on this subset of data}
\State \multiline{Apply the PCA transform built on this subset to the features in $S_j$ of the whole train set}
\State Replace the features $S_j$ in $D_i$ with the PCA features.
\EndFor
\State Build C4.5 Classifier $F_i$ on transformed data $D_i$.
\EndFor
\end{algorithmic}
\end{algorithm}

In summary, for each tree, rotation forest partitions the feature set, performs a restricted PCA on each of these subsets (via class and case sampling), then recombines the features over the whole train set. Sampling is performed independently on each feature subset for each tree, meaning it is a fundamentally different process to bagging or bootstrapping for the whole tree. The algorithm has three principle parameters: the number of trees $k$ (step 2); the number of features per set $f$ (step 3); and the proportion of cases to select ($p$) (step 7). A further potential parameter is the probability of selecting a class (step 6), but in the Weka implementation this is hard coded to 0.5. The other Weka defaults are $k=10$, $f=3$ and $p=0.5$. We explore the effect of these parameters on performance through sensitivity analysis in Section~\ref{section:sensitivity}. The three key design components where rotation forest differs from random forest are: the base classifier is C4.5 not RandomTree; all attributes are used for each base classifier rather than a sample; and the data are subsampled in groups then transformed with PCA. We investigate the importance of these components in an ablative study in Section~\ref{section:ablative}. There has been little experimental investigation into the performance of rotation forest. The inventors of rotation forest, Kuncheva and Rodriguez, investigate the impact of design features and parameters of rotation forest~\cite{kuncheva07experimental}. We discuss how this work relates to our investigation in  Sections~\ref{section:ablative} and~\ref{section:sensitivity}.  An empirical comparison of variants on rotation forest classifiers for customer churn prediction is presented in~\cite{debock11empricial}. Default values were used to assess performance on on four confidential datasets, the attributes of which are not explicitly described. In~\cite{liu08cancer} rotation forest was applied to two well known microarray cancer datasets (the breast cancer dataset and the prostate cancer dataset) and found to be more effective than bagging or boosting. An algorithm similar to rotation forest based on random rotations rather than PCA is described in~\cite{blaser16random}. They evaluated alternative algorithms on a subset of 29 UCI data with categorical and real-valued attributes with three forms of scaling. These contained both classification and regression problems. The criteria for dataset selection is not stated.  The R implementation of rotation forest differs from the Weka version (for example, instances are seemingly sampled by class at the group stage) and the results presented are not directly comparable to ours.

\section{Experimental design}
\label{section:experiments}
The UCI dataset archive\footnote{http://archive.ics.uci.edu/ml/index.php} is widely used in the machine learning literature. An extensive evaluation of 179 classifiers on 121 datasets from the UCI archive, including different implementations of notionally the same classifier, was performed by~\cite{delgado14hundreds}. Overall, they found that the random forest (RandF) algorithms maintained the highest average ranking, with support vector machines (SVM), neural networks and boosting achieving comparable performance. There was no algorithm significantly more accurate than all others on average. Although it has since been identified that the overlap between validation and test datasets may have introduced bias~\cite{Wainberg16randomforests}, these results mirror our own experience with these classifiers. To avoid any suspicion of cherry picking datasets, we initially carried out all our experiments with resamples of 121 datasets. These results are available on the accompanying website. However, many of these datasets are not ideal for a comparative study. Many are very small, and have been formatted in an unusual way: categorical attributes have been made real-valued in a way that may confound the classifiers. Based on reviewer feedback we have filtered the 121 datasets to select only those with what were originally continuous attributes and no missing values. We have removed datasets considered to be toy problems  (balloons, flags, led-display, lenses, monks-1, monks-2, monks-3, tic-tac-toe, trains, zoo annd wine). There are also several groups of strongly related datasets (acute*, breast-cancer*, cardio*, chess*, energy* heart*, monks*, musk* oocytes*, pit-bri*, plant*, spect*, vert-col-2*, waveform* and wine-quality*). We have selected either none (if attributes are categorical) or one of each of these (if all attributes are continuous). If we selected one, we took the problem with the highest number of attributes. Finally, image-segmentation and statlog-image are derived from the same base data. We removed statlog-image. This process leaves us with just 39 datasets. We stress that we did this selection without any consideration of the accuracy of alternative classifiers. The 121 datasets and associated results are available from the website.  A summary of the data is provided in Table~\ref{table:uciDatasets} in Appendix A.

The UCR archive is a collection of real-valued time series classification (TSC) univariate and multivariate datasets\footnote{http://www.timeseriesclassification.com}. A summary of the data is provided in Table~\ref{table:TSCproblems} in Appendix A and~\cite{bagnall18mtsc,dau18archive} provide an overview of the datasets. A recent study~\cite{bagnall17bakeoff} implemented 18 state-of-the-art TSC classifiers within a common framework and evaluated them on 85 datasets in the archive. One of the best approaches was the shapelet transform~\cite{hills14shapelet}. Shapelets are discriminatory subseries in the original dataset. The shapelet transform separates the finding of shapelets from the classification stage. Application of the shapelet transform selects $k$ good shapelets and creates a new dataset where each attribute represents the distance between a case and a shapelet. Hence, applying the transform creates a completely new classification problem. We denote this set of 85 shapelet transformed datasets as the ST-UCR datasets.

Experiments are conducted by averaging over 30 stratified resamples of data. For the UCI data, 50\% of the data are taken for training, 50\% for testing. The UCR archive provides a default train/test split. We perform stratified resamples using the number of train and test cases defined in these default splits, where the first resample is always the provided train/test split. Resample creation is deterministic and can be reproduced using the method \texttt{InstanceTools.resampleInstances}, or alternatively all resamples can be downloaded directly. The shapelet transform is performed independently on each resample. All code is available and open source\footnote{https://bitbucket.org/TonyBagnall/time-series-classification}. In the course of experiments we have generated gigabytes of prediction information and results. These are available in raw format from the correspondence author and in summary spreadsheets from the website.

We always compare classifiers on the same resamples. For comparing two classifiers on a single dataset we perform both a paired t-test and a sign rank test over the resamples. We note that this process inflates type I error, because the samples are not independent~\cite{demsar06comparisons}. Hence our main focus is comparing classifier averages taken over resamples on multiple datasets.  For comparing two classifiers over multiple datasets we take the average over all resamples of a single dataset and perform pairwise tests on the averages. For comparing multiple classifiers on multiple datasets, we follow the recommendation of Dem\v{s}ar~\cite{demsar06comparisons} and use the Friedmann test to determine if there were any statistically significant differences in the rankings of the classifiers.
However, following recent recommendations in \cite{benavoli16pairwise} and \cite{garcia08pairwise}, we have abandoned the Nemenyi post-hoc test originally used by~\cite{demsar06comparisons} to form cliques (groups of classifiers within which there is no significant difference in ranks). Instead, we compare all classifiers with pairwise Wilcoxon signed rank tests, and form cliques using the Holm correction (which adjusts family-wise error less conservatively than a Bonferonni adjustment).

We assess classifier performance by four statistics of the predictions and the probability estimates. Predictive power is assessed by test set error and balanced test set error. The quality of the probability estimates is measured with the negative log likelihood (NLL). The ability to rank predictions is estimated by the area under the receiver operator characteristic curve (AUC). For problems with two classes, we treat the minority class as a positive outcome. For multiclass problems, we calculate the AUC for each class and weight it by the class frequency in the train data, as recommended in~\cite{provost03pet}. All cross-validation uses ten folds.

\section{Comparison of classifiers}
\label{section:comparison}

\subsection{Comparison of Tuned Classifiers on UCI Datasets}

We compare a tuned support vector machines with RBF kernel (SVMRBF) and polynomial kernel (SVMP), a neural network with a single hidden layer using the Broyden-Fletcher-Goldfarb-Shanno method~\cite{Nocedal2006a} to optimise squared error (MLP1) and another MLP with up to two hidden layers (MLP2), a boosted tree ensemble (XGBoost), random forest (RandF), and rotation forest (RotF).

We use the Weka implementations for SVM (class SMO), MLP1 (MLPClassifier), MLP2 (MultiLayerPerceptron) RandF (RandomForest) and RotF (RotationForest). For gradient boosting we use the XGBoost implementation\footnote{https://github.com/dmlc/xgboost}. To get probability estimates, we used multi:softprob for XGBoost and logistic regression functions fitted to the output for SVM.

\begin{table*}[!htbp]
\caption{Tuning parameter ranges for SVMRBF, RandF, MLP, XGBoost and RotF. $c$ is the number of classes and $m$ the number of attributes}
\label{tab:ranges}
\begin{center}
\begin{tabular}{|llc|}\hline
Classifier & Parameter & Range \\ \hline
SVMP & Regularisation $C$ (33 values) & $\{2^{-16},2^{-15},\ldots,2^{16}\}$ \\
1188    & Exponent $p$ (6 values)& $\{1,2,\ldots,6\}$ \\
combinations   & Constant $b$ (6 values)& $\{0,1,\ldots,5\}$ \\
\hline
SVMRBF & Regularisation $C$ (33 values) & $\{2^{-16},2^{-15},\ldots,2^{16}\}$ \\
1089    & variance $\gamma$ (33 values)& $\{2^{-16},2^{-15},\ldots,2^{16}\}$ \\
combinations & &\\
\hline
Random Forest & number of trees (10 values) & $\{10,100,200,\ldots,900\}$ \\
1000			& feature subset size (10 values)& $\{\sqrt{m},(\log_2{m}+1),m/10,m/9,\ldots,m/3\}$ \\
combinations              & max tree depth (10 values)& $\{0,m/9,m/8\ldots,m\}$\\ \hline
Rotation Forest           & number of trees (10 values) & $\{10,100,200,\ldots,900\}$ \\
1000			          & attributes per group (10 values)& $\{3,4,\ldots,12\}$ \\
combinations              & sampling proportion (10 values)& $\{0.1,0.2,\ldots,1.0\}$\\
            \hline
MLP1    & \# hidden units (33 values)	& \{min($c$,$m$), ..., $c+m$\} \\
1089	& ridge factor (33 values)			& \{$log_{10}(-4)$, ..., $log_{10}(-1)$\}  \\
combinations & & \\\hline
MLP2 & hidden layers (2 values) & $\{1,2\}$\\
	& nodes per layer (4 values) & $\{c,m,m+c,(m+c)/2\}$ \\
    & learning rate (8 values)	& $\{1,1/2,1/4,\ldots,1/(2^7)\}$\\
1024	& momentum (8 values)			& $\{0,1/8,2/8,\ldots,7/8\}$\\
combinations	& decay	(2 values)		& $\{true,false\} $ \\ \hline
XGBoost 	&  number of trees (10 values)& $\{ 10, 25, 50, 100, 250, 500, 750, 1000, 1250, 1500 \}$ \\
1000			& learning rate (10 values)	&  $\{10^{-6}, 10^{-5}, 10^{-4}, 0.01, 0.05, 0.1, 0.15, 0.2, 0.25, 0.3\}$   \\
combinations				& max tree depth (10 values)	& $\{ 1,2,3,4,5,6,7,8,9,10 \}$ \\
\hline
\end{tabular}
 \end{center}
\end{table*}

\begin{figure*}[!ht]
	\centering
\begin{tabular}{cc}
       \includegraphics[width =8cm, trim={2cm 6.5cm 1cm 3cm},clip]{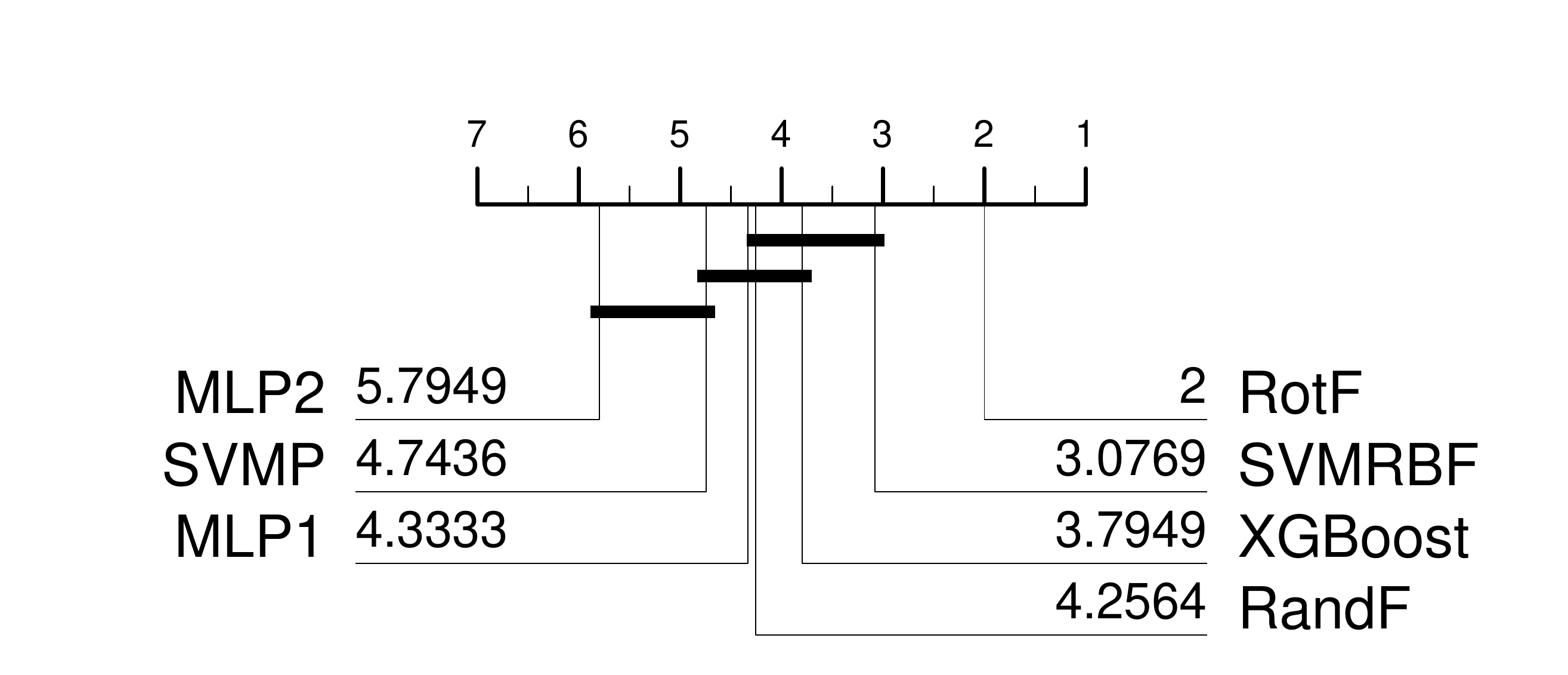}              	
&
       \includegraphics[width =8cm, trim={2cm 6.5cm 1cm 3cm},clip]{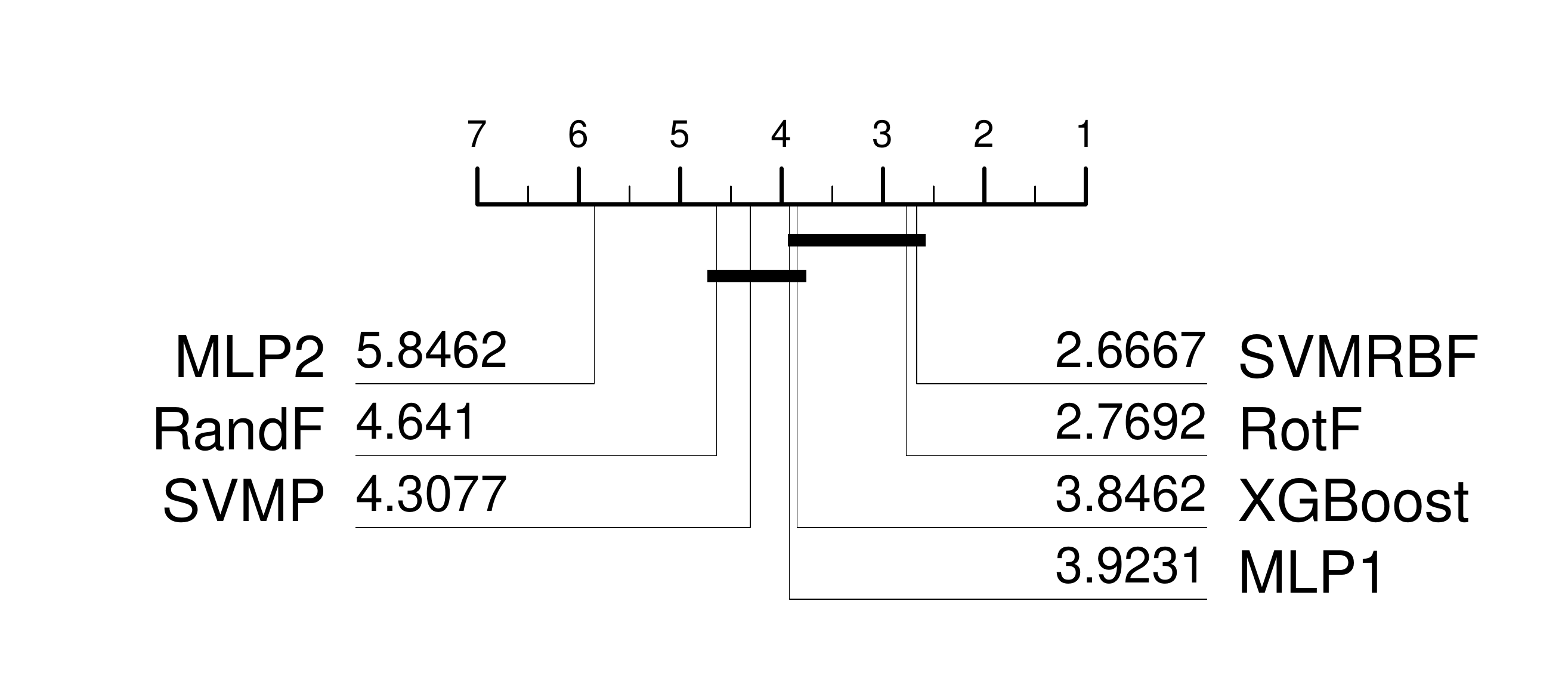}  \\
(a) Error & (b) Balanced Error \\
       \includegraphics[width =8cm, trim={2cm 6.5cm 1cm 3cm},clip]{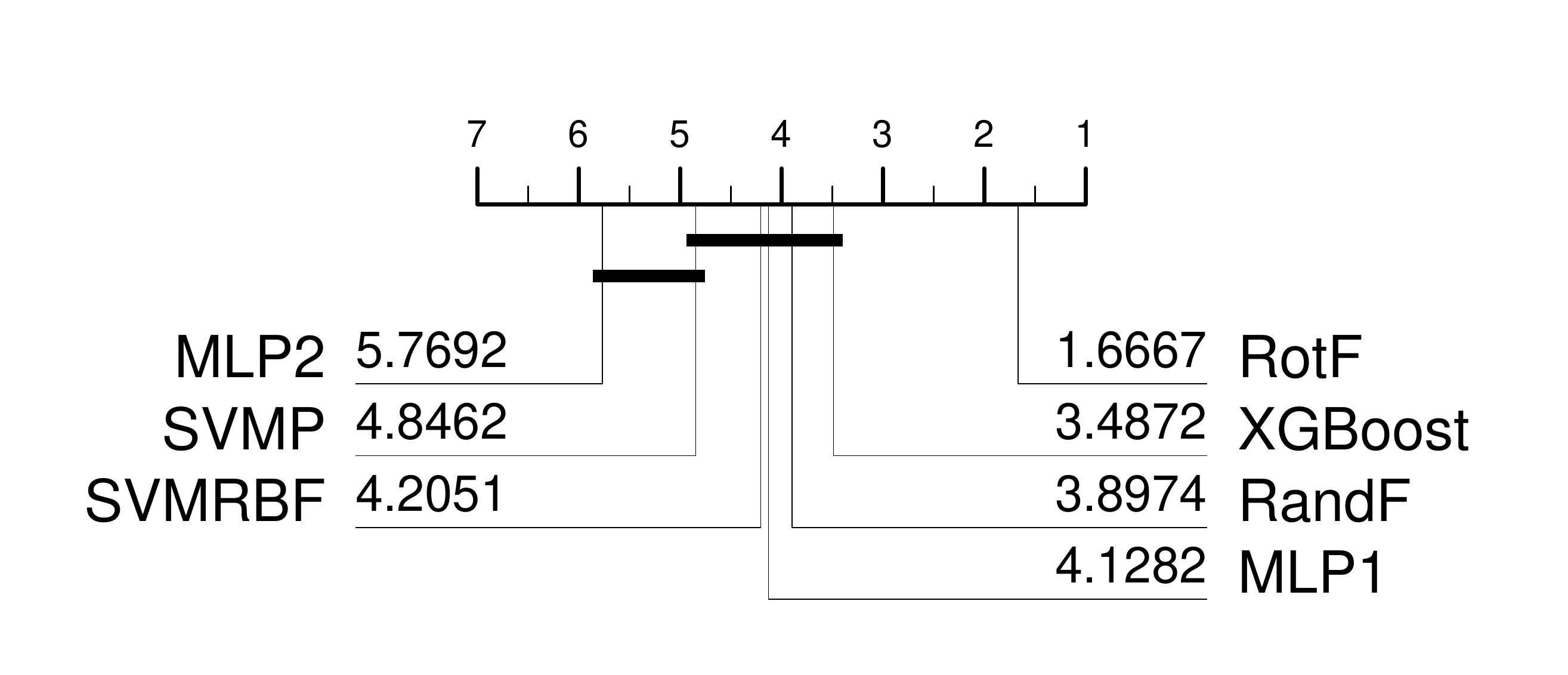}
&
       \includegraphics[width =8cm, trim={2cm 6.5cm 1cm 3cm},clip]{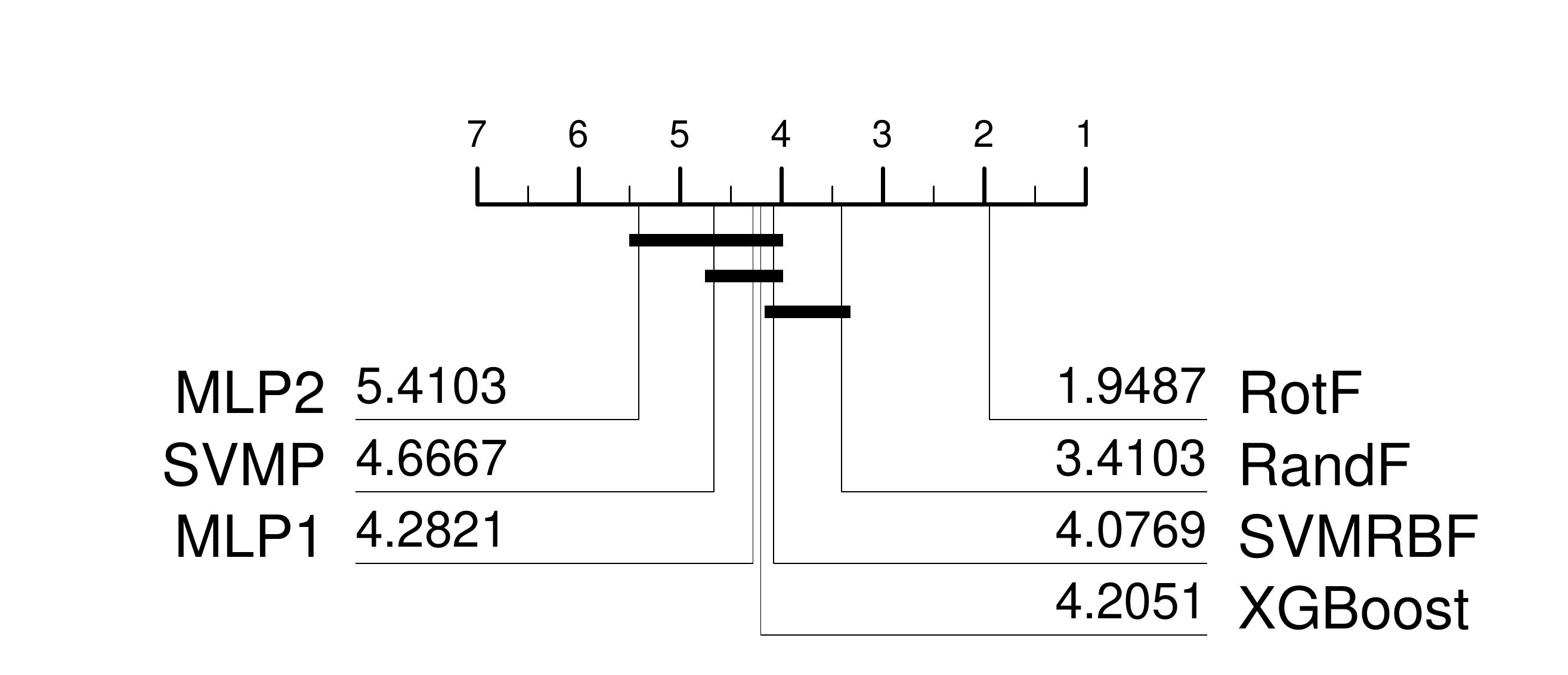}              	\\
(c) AUC & (d) NLL \\          
       \end{tabular}
       \caption{Critical difference diagrams for tuned classifiers on 39 UCI data. The classifiers are: Multi layer perceptron with a single hidden layer (MLP1) and either one or two hidden layers (MLP2); support vector machine with polynomial (SVMP) and radial basis function (SVMRBF) kernels;  extreme gradient boosting (XGBoost); random forest (RandF); and rotation forest (RotF). Parameter ranges searched for are given in Table~\ref{tab:ranges}.}
       \label{fig:tuned}
\end{figure*}

Table~\ref{tab:ranges} shows the parameter combinations evaluated for each classifier. On each resample, every combination is evaluated on the train data using a ten fold cross-validation. The combination with the lowest cross-validated error on the train data is selected, and a model is constructed with these values using all the train data. It is then evaluated on the test data. This means we construct approximately 300,000 models for each dataset and classifier combination (30 resamples, 1000 model combinations and 10 fold cross-validation for each), i.e. over 10 million models overall for just the 39 UCI problems (and approximately 60 million on all datasets).

The results, shown in Figure~\ref{fig:tuned}, demonstrate that rotation forest is significantly better than the other six tuned classifiers in terms of error, AUC and NLL. There is no significant difference between SVMRBF, RandF and XGBoost.   RandF, XGBoost, SVMP and MLP1 also form a clique for error.  The criteria where RotF has the biggest advantage is AUC. The only criteria where RotF is not significantly better than the alternatives is balanced error.

These results lend support to our core hypothesis that rotation forest is, on average, significantly better than the other classifiers we have compared it to, even when extensive tuning is performed.

\subsection{Comparison of Tuned Classifiers on UCR and ST-UCR Datasets}
\label{section:ucr}
Rotation forest is clearly better on average than the competitors on the UCI data. However, a reliance on one set of datasets can cause an over interpretation of results.  The real question is whether this generalises to other problems which meet our criteria of having all real-valued attributes. To test this, we evaluate classifiers on time series classification problems from the UCR archive and data derived from the archive by shapelet transform. We abandon MLP1 and MLP2 for practical reasons: they just take too long to run on UCR and ST-UCR data when tuned. The results from UCI (Figure~\ref{fig:tuned}) suggest MLP1 and MLP2 are unlikely to perform as well as rotation forest. We have verified this on the smaller problems on which we were able to run MLP1 and MLP2, but exclude the results for brevity. Furthermore, SVMRBF, SVMP and  XGBoost (in particular) were very slow on some of the problems. Hence, we present UCR and ST-UCR results on 78 of the 85 datasets in the archive. We stress that the datasets that were excluded were done for computational reasons, not through any cherry picking process. A full list of datasets and results are available on the accompanying website.

\begin{figure*}[!ht]
	\centering
\begin{tabular}{cc}
       \includegraphics[width =8cm, trim={2cm 6.5cm 1cm 3cm},clip]{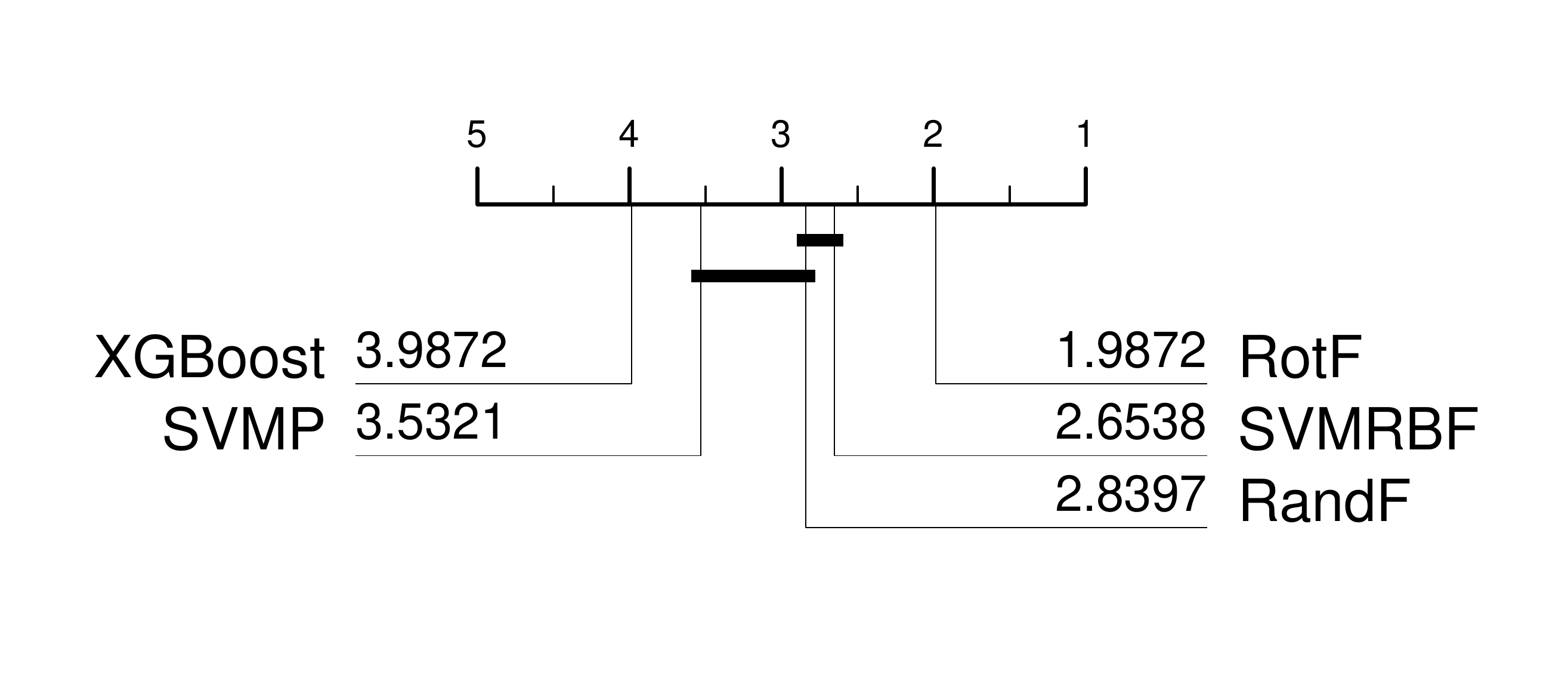}              	
&
       \includegraphics[width =8cm, trim={2cm 6.5cm 1cm 3cm},clip]{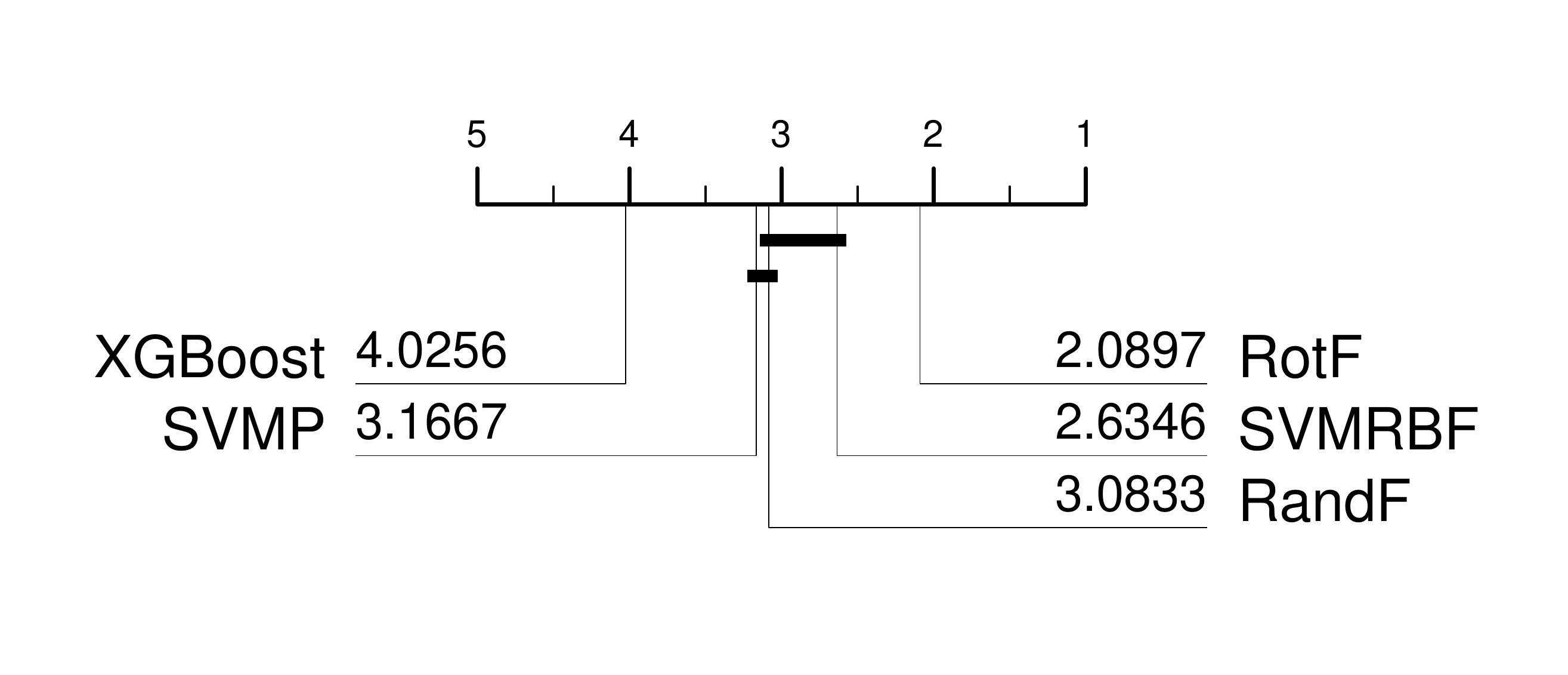}  \\
(a) Error & (b) Balanced Error \\
       \includegraphics[width =8cm, trim={2cm 6.5cm 1cm 3cm},clip]{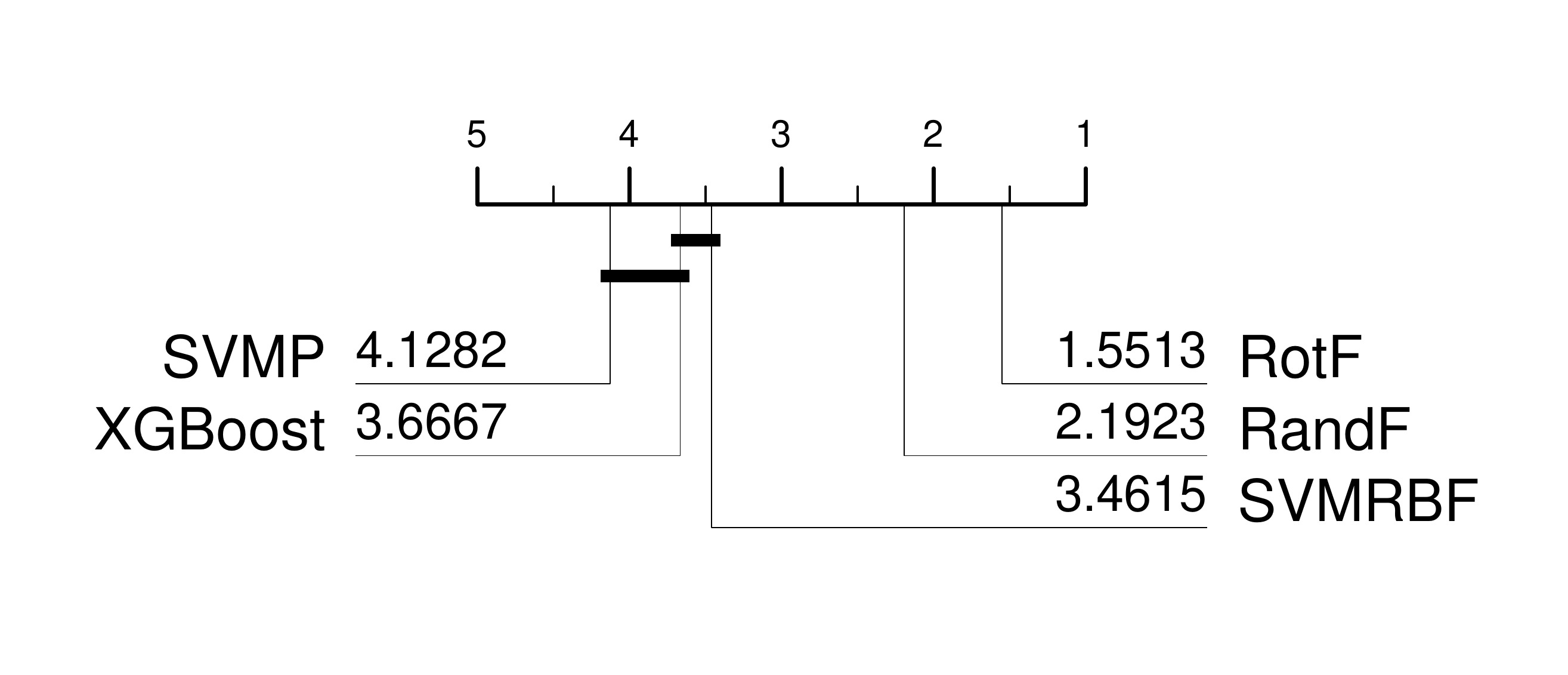}
&
       \includegraphics[width =8cm, trim={2cm 6.5cm 1cm 3cm},clip]{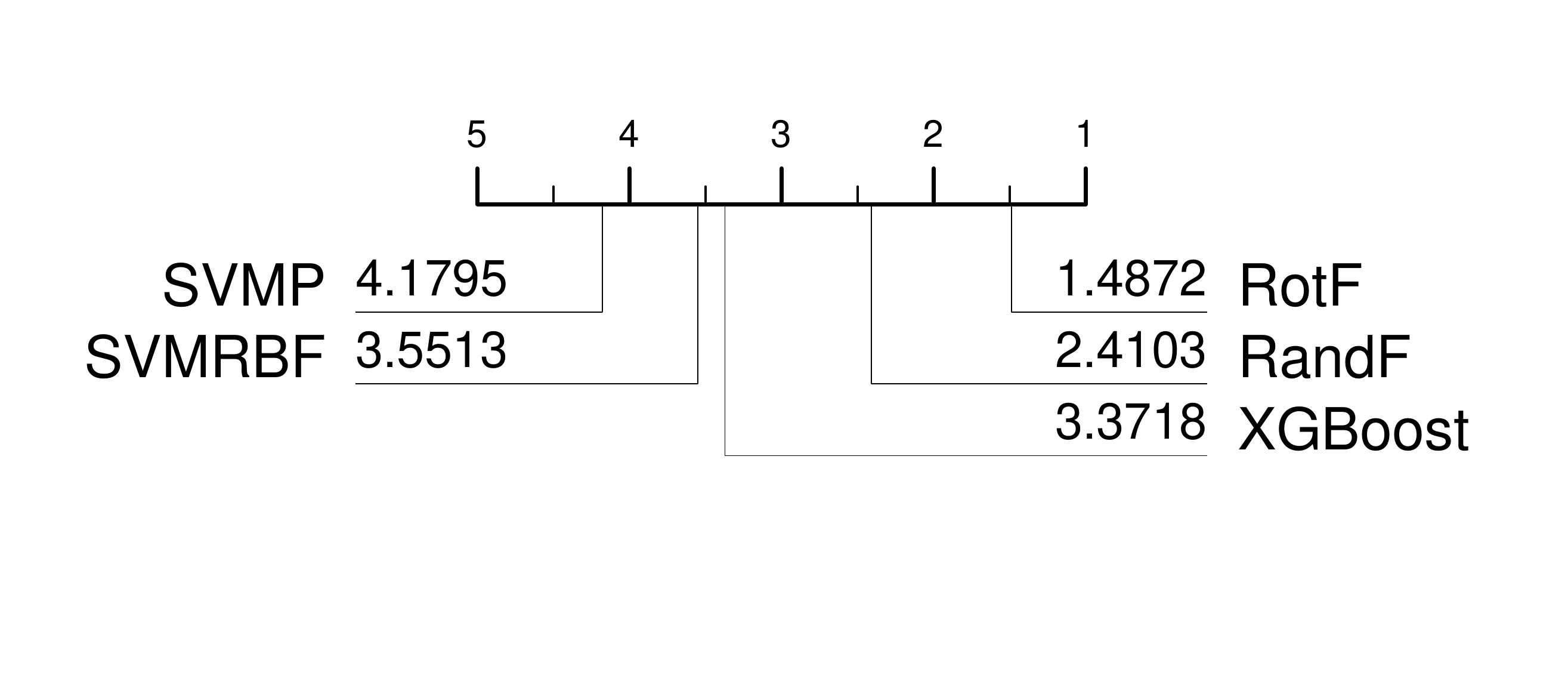}              	\\
(c) AUC & (d) NLL \\          
       \end{tabular}
       \caption{Critical difference diagrams for tuned classifiers on 78 UCR data. Parameter ranges searched for are given in Table~\ref{tab:ranges}. Ranks are formed from averages over 30 resamples on each problem.}
       \label{fig:ucr1}
\end{figure*}
Figure~\ref{fig:ucr1} shows the critical difference diagrams for 78 UCR data. The results closely mirror those on the UCI data. Rotation forest is significantly better than the other classifiers under all four metrics. It is most demonstrably better at ranking cases (AUC) and producing probability estimates. It is not as relatively effective when considering balanced error, but it is still significantly better. There is no significant difference in terms of error between SVMRBF and RandF, but RandF is ranked more highly than the other three classifiers on AUC and NLL.

\begin{figure*}[!ht]
	\centering
\begin{tabular}{cc}
       \includegraphics[width =8cm, trim={2cm 6.5cm 1cm 3cm},clip]{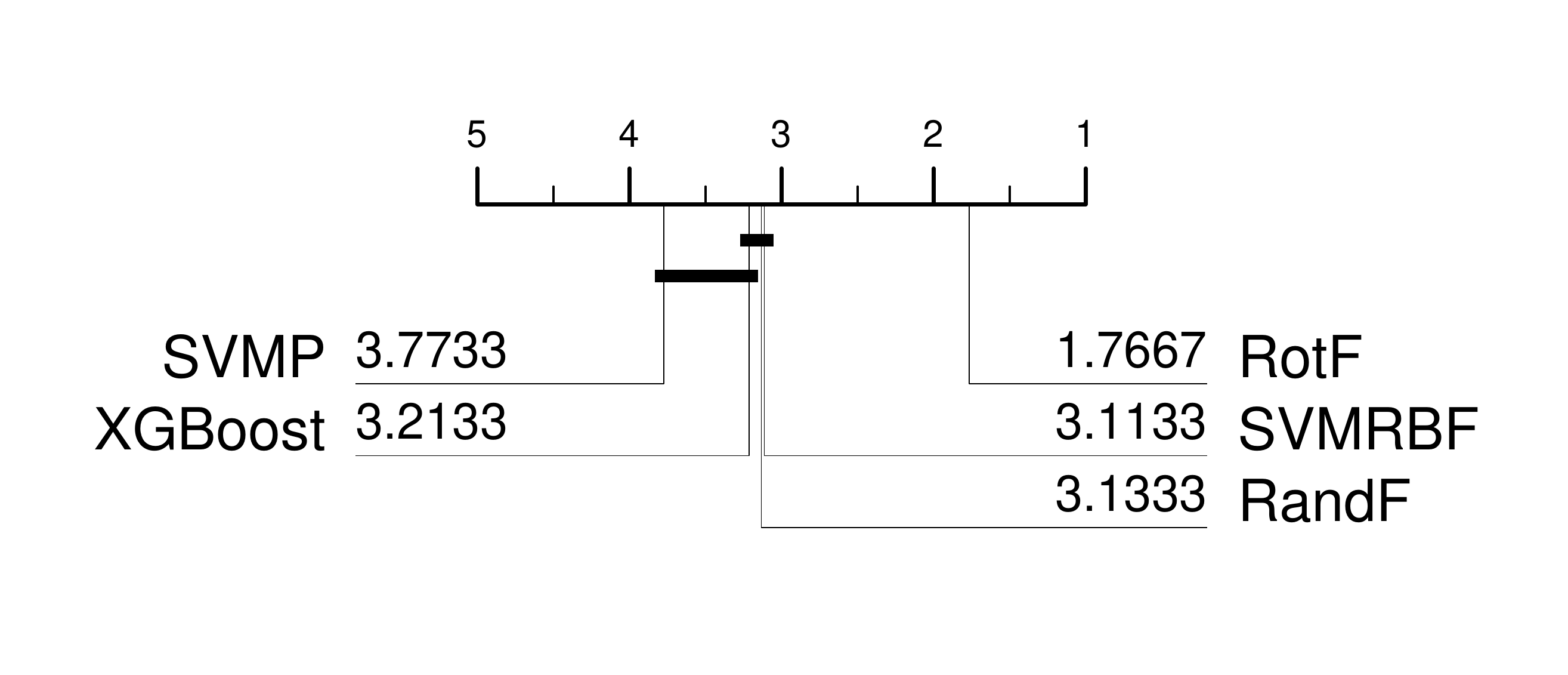}              	
&
       \includegraphics[width =8cm, trim={2cm 6.5cm 1cm 3cm},clip]{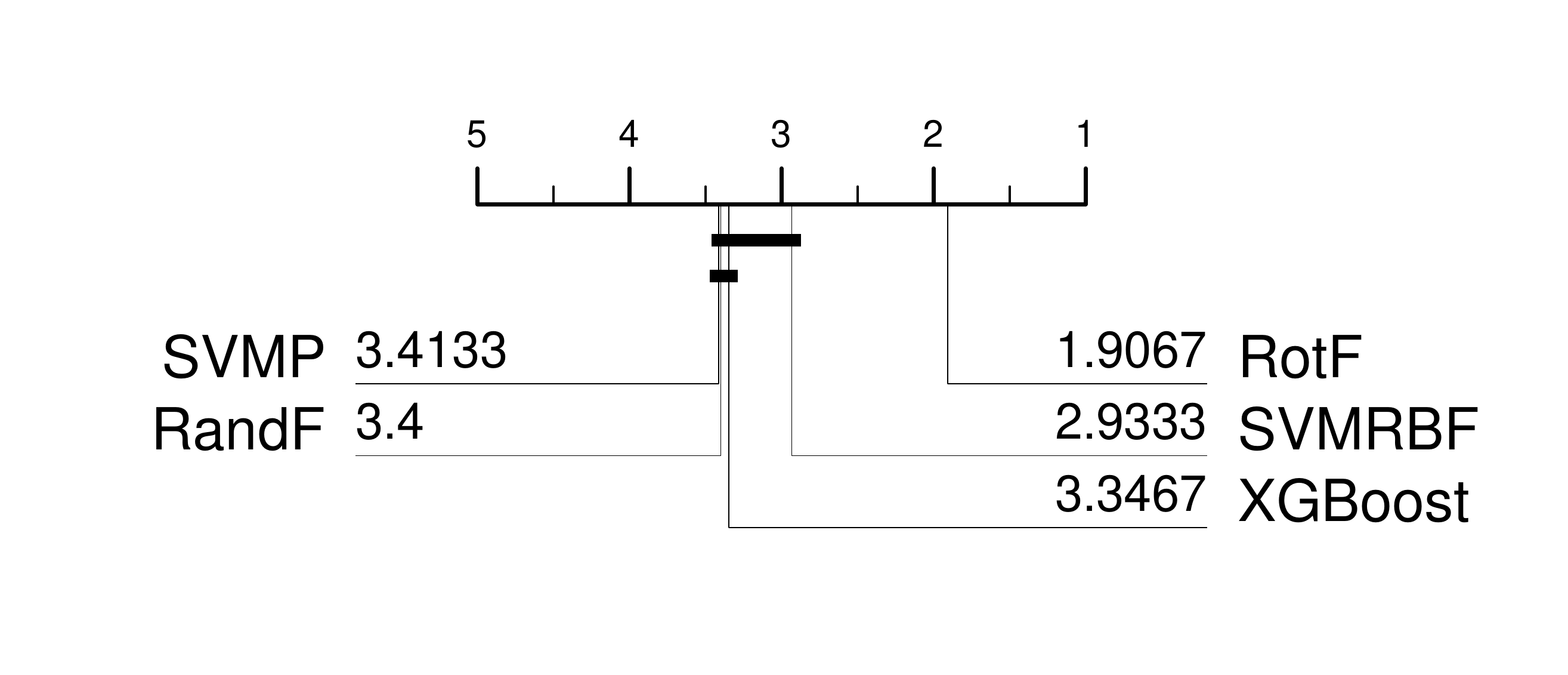}  \\
(a) Error & (b) Balanced Error \\
       \includegraphics[width =8cm, trim={2cm 6.5cm 1cm 3cm},clip]{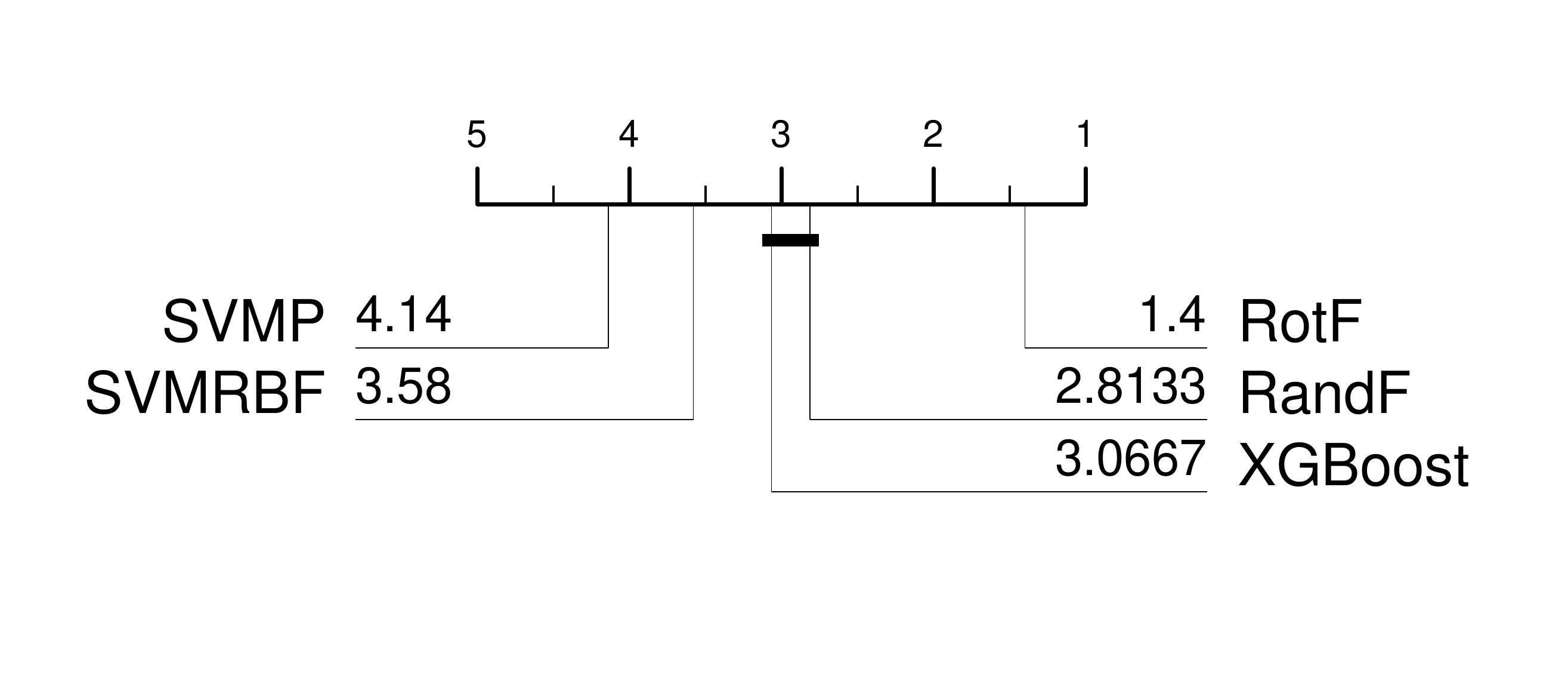}
&
       \includegraphics[width =8cm, trim={2cm 6.5cm 1cm 3cm},clip]{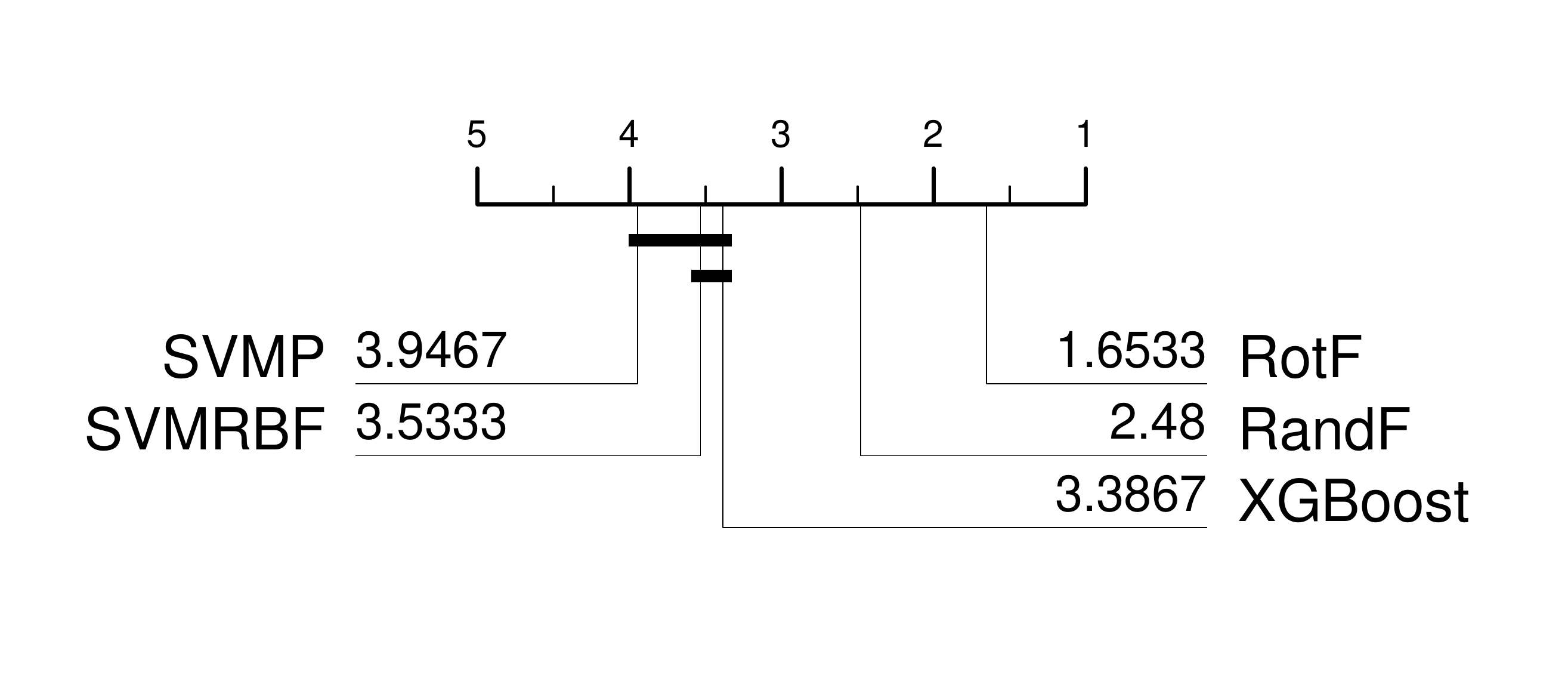}              	\\
(c) AUC & (d) NLL \\          
       \end{tabular}
       \caption{Critical difference diagrams for tuned classifiers on 78 ST-UCR data. Parameter ranges searched for are given in Table~\ref{tab:ranges}. }
       \label{fig:st-ucr}
\end{figure*}
Figure~\ref{fig:st-ucr} shows the critical difference diagrams for 78 ST-UCR data. The pattern of performance is very similar to that observed on UCI and UCR. Rotation forest is the best overall with all four measures. The performance advantage of rotation forest is even more pronounced than with other experiments. We think this is because the shapelet attributes are often highly correlated, and the rotation at the heart of rotation forest is best able to compensate for this.

Given the time it takes to tune, it is worthwhile asking whether it makes a significant difference on average. Surprisingly, tuning makes no significant improvement to rotation forest or random forest, as long as reasonable default values are used. The classifier to gain significantly from tuning is SVMRBF, where the improvement is dramatic. Tuning makes the classifier over 3\% more accurate on average, and gives a significant improvement to SVMRBF on over half of the problems (when comparing over resamples).  Given these results, we continue the comparative study using random forest and rotation forest with fixed parameters (given in Table~\ref{tab:fixed}). This seems reasonable, since the two classifiers have a similar basic structure, and tuning carries huge computational cost.
\begin{table}
\caption{Fixed parameters for random forest and rotation forest. $m$ is the number of attributes. Any parameter not stated is set to the default value in the implementation used for experiments. }
\label{tab:fixed}
\begin{center}
\begin{tabular}{|llc|}\hline
Classifier & Parameter & Value \\ \hline \hline
Random Forest & number of trees & 500  \\
			& feature subset size & $\sqrt{m}$ \\
              & max tree depth & no limit\\ \hline
Rotation Forest & number of trees ($k$) & 200 \\
			    & attributes per group ($f$)& 3\\
                & sampling proportion   ($p$)& 0.5 \\

                \hline

\end{tabular}
 \end{center}
\end{table}

\subsection{Comparison of random forest and rotation forest}

There are three objectives for the experiments described in the rest of this section.  Firstly, we perform a more detailed comparison of random forest and rotation forest. Secondly, we assess how well classifiers that make no explicit use of the time information in the UCR classification problems compare to the most widely used benchmark algorithm for TSC, dynamic time warping. This is a sanity check to demonstrate that observed differences are not simply an artifact of choosing between two classifiers that are both unsuited to the type of problem.  Finally, we wish to evaluate whether using rotation forest on data transformed from the UCR archive can make a better bespoke TSC algorithm, and compare this to state of the art for bespoke TSC algorithms. This serves to provide evidence of the importance of classifier selection.

\subsubsection{Random forest vs rotation forest on UCR and ST-UCR data}

A comparison of random forest and rotation forest on the UCR and ST-UCR datasets is shown in Table~\ref{tab:stucr1}. The first two rows show the p-values for the Wilcoxon sign rank test for equality of median and the paired t-test for equality of mean. We can reject the null hypotheses of equality of averages on both sets of data. The mean difference and win/draw/loss counts are presented for information only. The last row displays the number of problems where there is a significant difference between the paired resamples.
\begin{table}[!htb]
\caption{A comparison of rotation forest and random forest on the UCR and ST-UCR datasets. The pairwise tests compare the average over 30 resamples for each dataset. The wins refer to when rotation forest is better than random forest on average. The sig wins are the counts of when rotation forest is significantly better than random forest on a specific dataset, as measured over the 30 resamples.}
\label{tab:stucr1}
\begin{center}
\begin{tabular}{|l|l|c|}\hline
Classifier      & UCR 		& ST-UCR \\ \hline
MW-test p-value			& 0.0		&	0.011	\\
Paired t-test p-value 	& 0.0		&	0.039	\\
Mean difference 		& 2.9\%		&	0.3\%	\\
Wins/draws/losses		&   70/1/14 & 	52/0/33   \\
Sig wins/draws/losses	& 54/28/3	&  15/64/6 \\
\hline
\end{tabular}
 \end{center}
\end{table}

\begin{figure}[!ht]
	\centering
\begin{tabular}{c}
       \includegraphics[width =8cm, trim={3cm 1cm 2cm 2cm},clip]{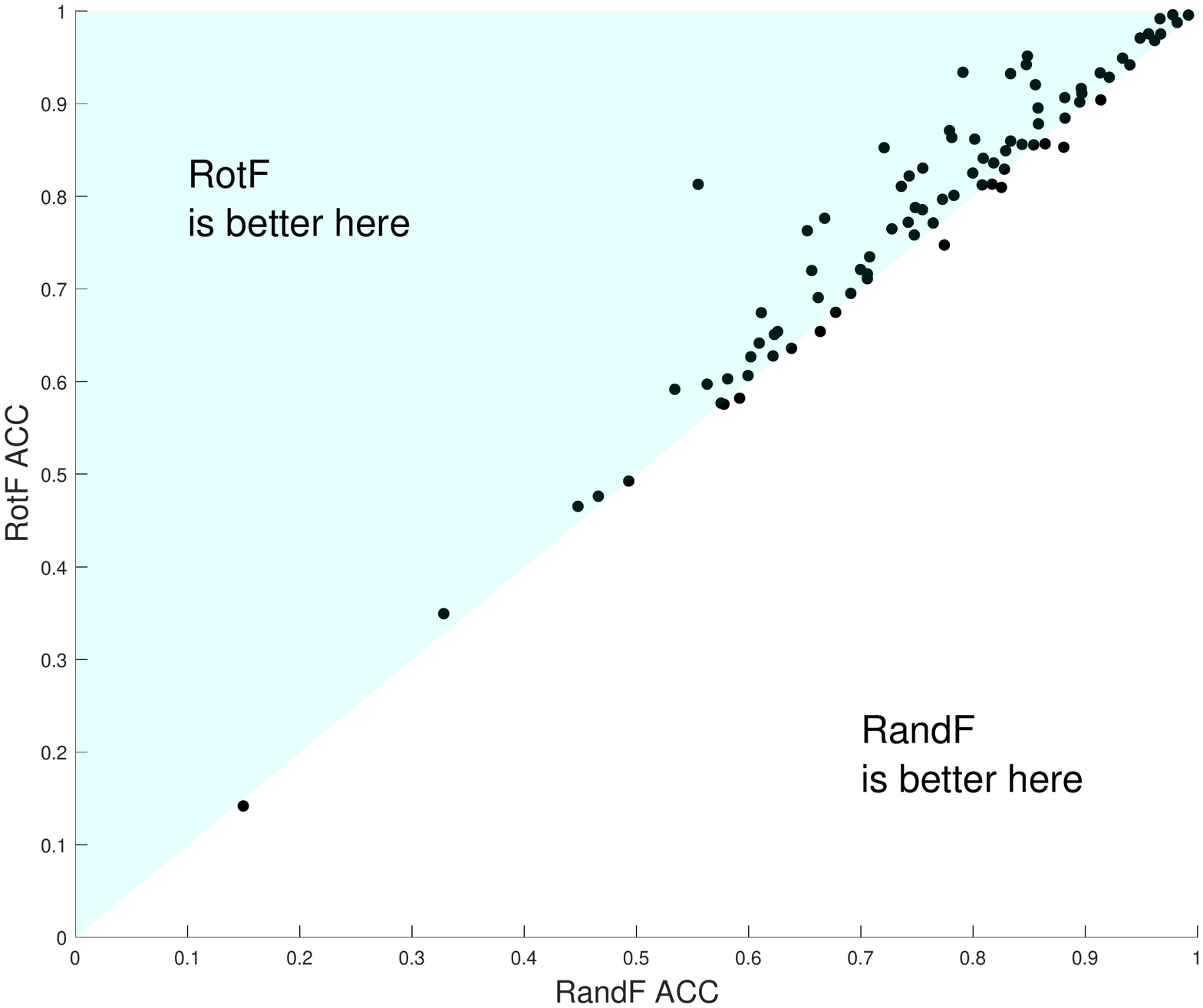}   \\
(a) UCR \\       
       \includegraphics[width =8cm, trim={3cm 1cm 2cm 2cm},clip]{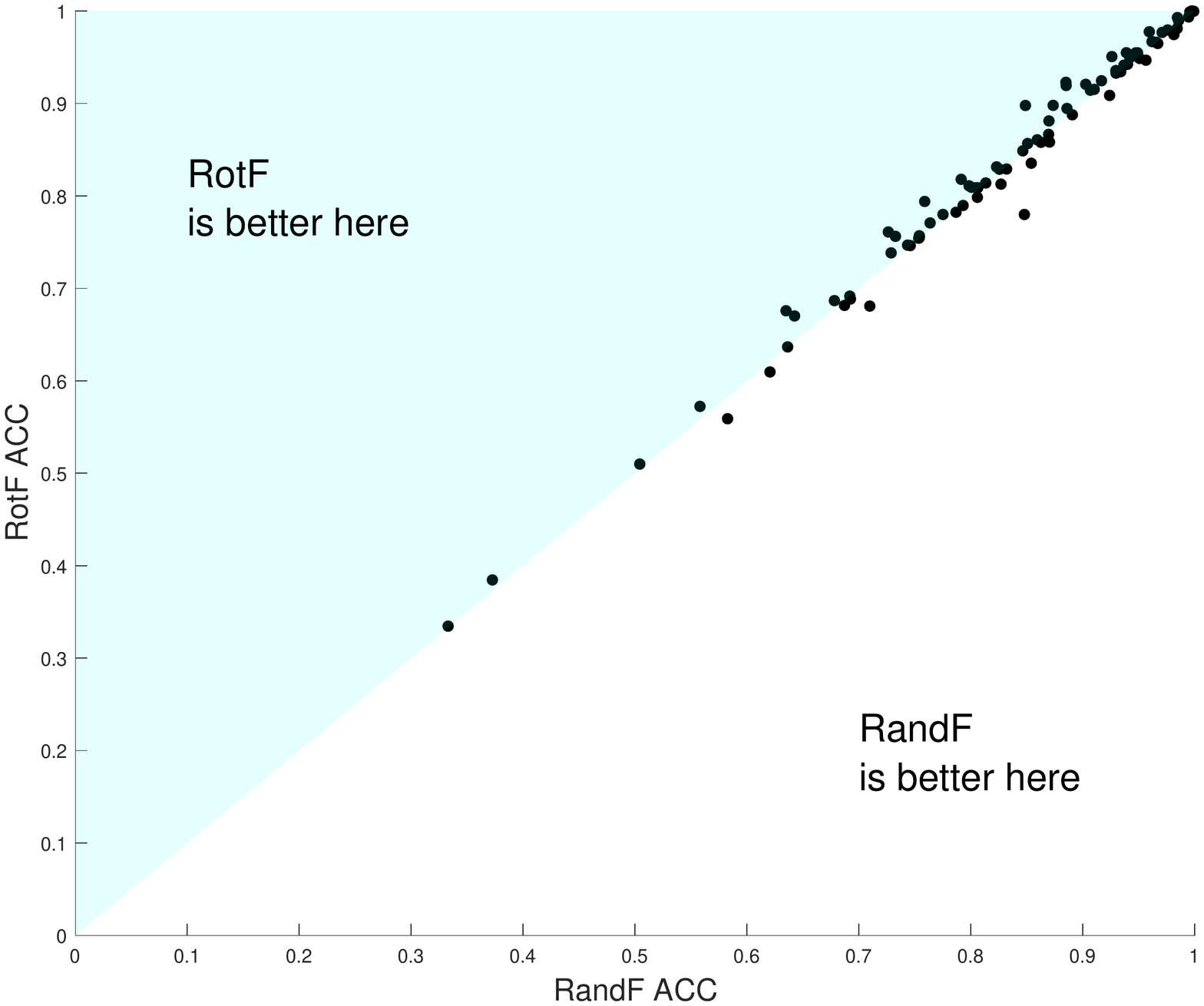}  \\
 (b) ST-UCR
       \end{tabular}
       \caption{Scatter plots for accuracies of random forest and rotation forest on two sets of data.}
       \label{fig:scatter}
\end{figure}

Rotation forest is significantly better than random forest on both sets of datasets using both tests. The difference is more marked for the UCR dataset, reflecting rotation forests ability to find underlying discriminatory auto-correlations in time series data. The difference is smaller on the ST-UCR data, where the ordering of attributes is irrelevant, but still significant. Figure~\ref{fig:scatter} serves to help visualise the differences between the classifiers by showing the scatter plots of test accuracies.
A comparison of balanced error, AUC and NLL yields a very similar pattern of results.

\subsubsection{Rotation forest vs dynamic time warping}

Rotation forest is not designed to handle time series data, so it is of interest to see how well it does in comparison to algorithms designed specifically for TSC. Dynamic time warping (DTW) distance with a 1 nearest neighbour classifier is considered a strong benchmark for TSC problems. DTW compensates for potential phase shift between classifiers through realignment that minimises the distance within constraints~\cite{ratanamahatana05threemyths}. The degree of warping allowed is determined by a warping window. We denote 1-NN dynamic time warping with full warping window as DTW. Commonly, DTW window size is set through cross-validation (DTWCV). The case for  DTW and DTWCV as benchmarks is commonly made. For example, recent papers have stated that {\em ``Many studies have shown that the One Nearest Neighbour Search with DTW (NN-DTW) outperforms most other algorithms when tested on the benchmark datasets"}\cite{tan18dtwcv} and {\em ``Over the last decade, the time series research community seems to have come to the consensus that DTW is a difficult-to-beat baseline for many time series mining tasks"}~\cite{dau18optimizing}. A comparative study found DTW and DTWCV were not significantly worse than many recently proposed bespoke TSC algorithms~\cite{bagnall17bakeoff}. As part of our goal in assessing the quality of rotation forest as a classifier, it is of interest to assess how well it performs on raw TSC data in comparison to DTW.
Because DTW and DTWCV are 1-NN classifiers they do not produce probability estimates nor rank the test cases, so there is no value in using NLL and AUROC as a comparison. We restrict our attention to comparing error.
\begin{figure}[!ht]
	\centering
       \includegraphics[width =8cm, trim={2cm 8cm 1cm 5cm},clip]{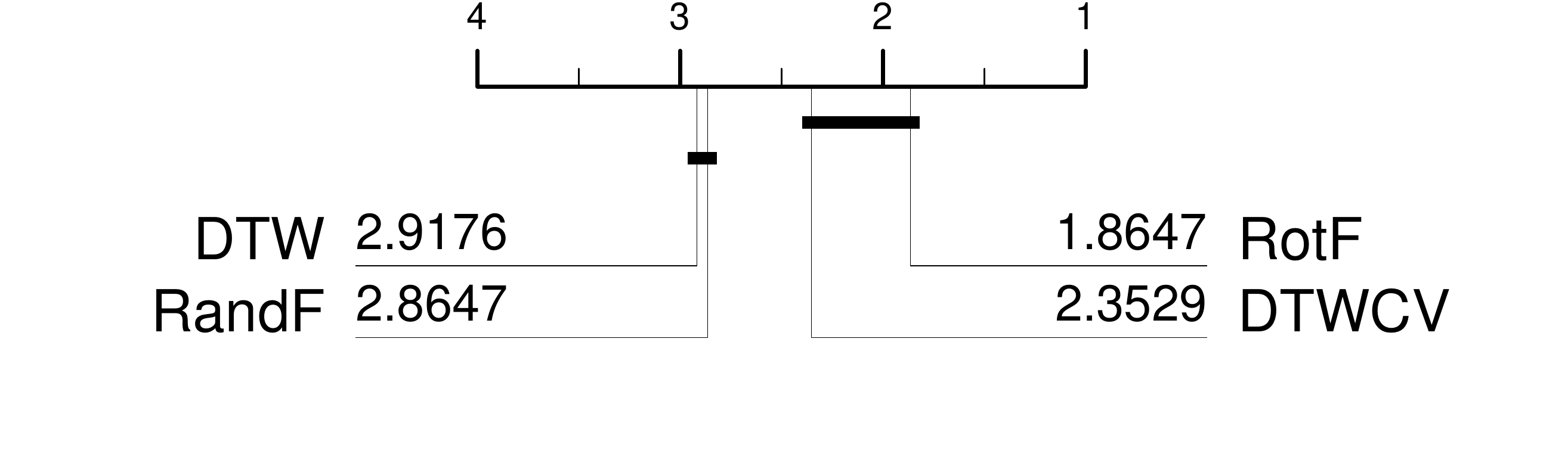}              	
       \caption{Critical difference diagram for the error of four classifiers on 85 UCR data. The classifiers are: random forest (RandF); rotation forest (RotF) full window dynamic time warping (DTW) and dynamic time warping with window size set through cross-validation (DTWCV).}
       \label{fig:ucr2}
\end{figure}
Figure~\ref{fig:ucr2} compares the forest results to the DTW errors. It shows that random forest is not significantly worse than DTW but is less accurate than DTWCV. Rotation forest, however, is significantly better than DTW and not significantly worse than DTWCV. Rotation forest beats DTWCV on 51 of the 85 problems and has on average 0.05\% lower error. These results demonstrate rotation forest's capacity to find discriminatory features in the auto-correlation and reinforce its general utility as a benchmark classifier. We conclude that rotation forest is the best standard classifier benchmark for TSC and is at least as good as DTWCV which is widely held to be the gold standard benchmark.

\subsubsection{Rotation forest against state-of-the-art TSC algorithms}

In the previous section we used the ST-UCR data to demonstrate that rotation forest is significantly better than random forest on a dataset unrelated to the UCI and without time dependencies. It also worthwhile placing the results in the context of alternative TSC algorithms to examine whether using rotation forest instead of an alternative algorithm makes a significant difference in comparison to other bespoke approaches. The most accurate classifier on average on the UCR data is the meta ensemble HIVE-COTE. This contains classifiers built on five alternative representations of the data:  time series forest (TSF)~\cite{deng13forest} is constructed on summary features of random intervals; the elastic ensemble (EE)~\cite{lines15elastic} is an ensemble of nearest neighbour classifiers using a range of alternative distance measures; random interval spectral ensemble (RISE)~\cite{lines18hive} is an ensemble of classifiers built on spectral transformations of random intervals; bag of symbolic-Fourier-approximation symbols (BOSS)~\cite{schafer15boss} is an ensemble constructed on histogram counts of repeating patterns in the data; and the shapelet transform ensemble (ST)~\cite{bostrom17binary} is an ensemble built on shapelet-transformed data (i.e. ST-UCR).
\begin{figure}[!ht]
	\centering
       \includegraphics[width =8cm, trim={5cm 13cm 4cm 11cm},clip]{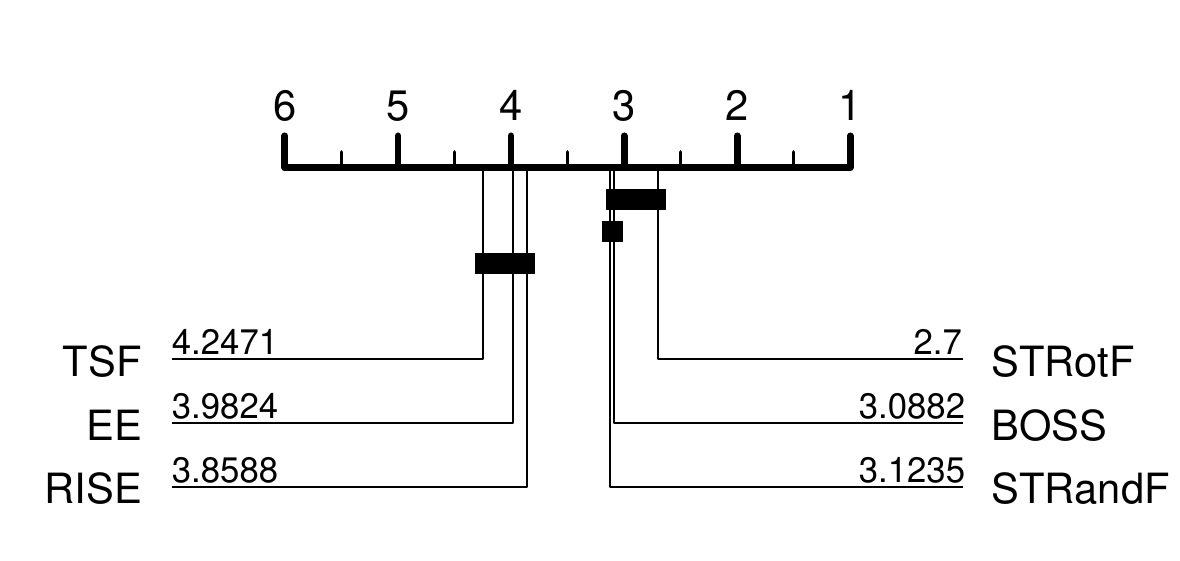}              	
       \caption{Critical difference diagram for the error of six bespoke time series classifiers on 85 UCR data. The classifiers are: shapelet transform~\cite{hills14shapelet}  with rotation forest (STRotF) and with random forest (STRandF); shapelet transform with rotation forest (STRotF); time series forest (TSF)~\cite{deng13forest}; elastic ensemble (EE)~\cite{lines15elastic}; random interval spectral ensemble (RISE)~\cite{lines18hive}; bag of symbolic Fourier approximation symbols (BOSS)~\cite{schafer15boss}.}
       \label{fig:shapelets}
\end{figure}
Figure~\ref{fig:shapelets} shows the relative performance of these four classifiers and the two forests constructed on the shapelet transform (STRandF and STRotF). The top clique consists of STRotF and BOSS. The second clique contains BOSS and STRandF. STRotF is significantly better than STRandF. Changing from random forest to rotation forest for the shapelet data not only provides significant reduction in error, but also makes the shapelet component of COTE the highest ranked overall. This demonstrates that a single simple design decision such as a change in base classifier can produce significant improvement. This leads us to the question of why is rotation forest better than random forest?






\section{From random forest to rotation forest: an ablative study}
\label{section:ablative}

To try to understand why rotation forest does so well, we deconstruct the algorithms to identify the differences. This work extends that conducted by the inventors of rotation forest, Kuncheva and Rodriguez~\cite{kuncheva07experimental}. They performed a similar study, where they assessed the importance of splitting the features and using PCA on a sparse matrix to perform the rotation. We consider an alternative question. Why does rotation forest perform better than random forest? In Section~\ref{section:rotf} we identified the two key differences between random forest and rotation forest:

\noindent  {\bf The sampling and transformation}. random forest uses bagging, i.e. it samples cases for each tree. Rotation forest samples cases for groups of attributes independently. Rotation forest then performs a PCA on the resulting groups. The PCA transform is obviously a key component of rotation forest. However, the way it is applied is not standard. It essentially uses a sparse transformation matrix. We compare three approaches to sampling and transformation: the random forest approach of bagging for each tree with no transformation (BAG); bagging followed by PCA on the entire data set (BAG+PCA); and the rotation forest mechanism of case sampling and PCA on a sparse transformation matrix (PCA).

\noindent {\bf The base classifier}: random forest uses random tree whilst rotation forest uses C4.5. We denote these factors RT and C4.5. In Weka, both these classifiers are based on selecting branching attributes based on information gain. The main difference between these classifiers is that RT randomly selects a subset of attributes at each node to evaluate as possible splitting criteria (we use $\sqrt{m}$ attributes).  To avoid over complicating the analysis, we compare using either RT or C4.5 with the default settings.

\begin{table}[!htp]
\centering
\begin{tabular}{llll}
Combination & Base Classifier & Transformation & Average Accuracy    \\ \hline
1 RandF     & RT              & BAG            & 85.07\%    \\
2           & RT              & BAG+PCA        & 87.12\%    \\
3           & RT              & PCA            & 87.37\%  \\
4           & C4.5            & BAG            & 80.49\%  \\
5           & C4.5            & BAG+PCA        & 86.82\%  \\
6   RotF    & C4.5            & PCA            & 87.16\%
\end{tabular}
\caption{All factor combinations for 6 hybrid classifiers, with the average accuracy and the rank on the 39 UCI datasets listed in Table~\ref{table:uciDatasets}. All combinations are trained with 200 trees.}
\label{tab:HybridComb}
\end{table}

  \begin{figure}[!ht]
	\centering
       \includegraphics[width =8cm, trim={0cm, 6cm, 0cm, 5cm}, clip]{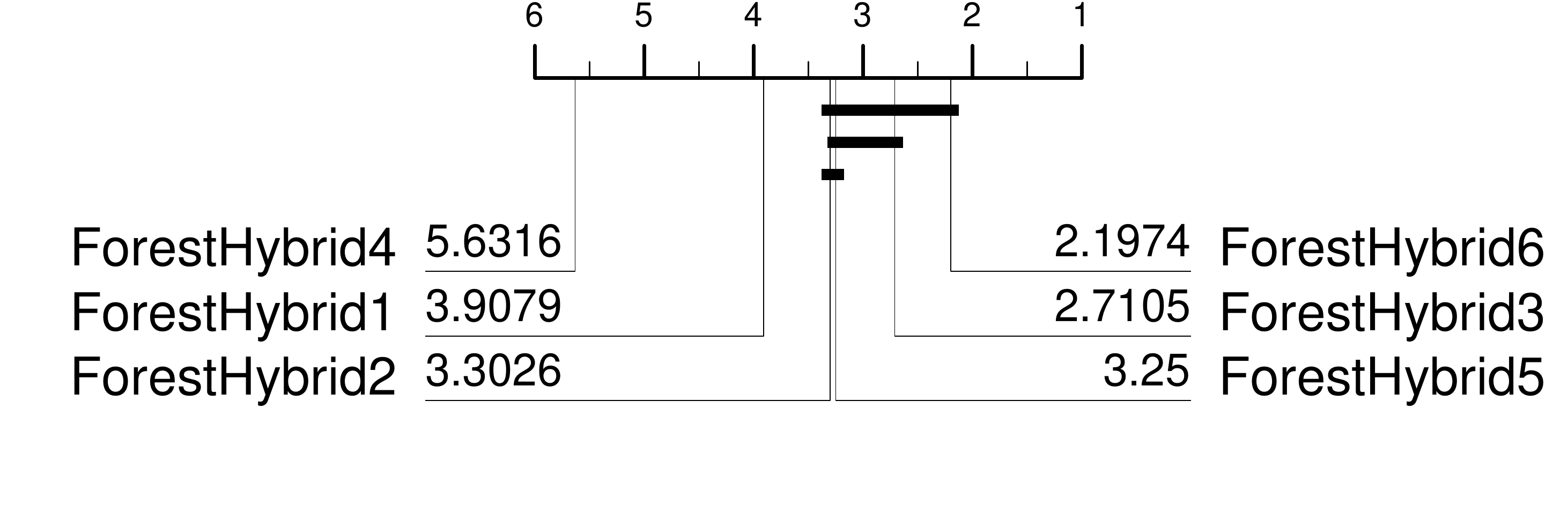}
       \caption{Critical difference diagram showing the rank of each parameter permutation between random forest and rotation forest, given the structural combinations listed in Table~\ref{tab:HybridComb}. Combination 1 is random forest, combination 6 rotation forest.}
       \label{fig:HybridCD}
\end{figure}

Table~\ref{tab:HybridComb} summarises the results for the 6  combinations, whilst Figure~\ref{fig:HybridCD} shows the resulting critical difference diagram. Rotation forest is highest ranked, but it not significantly better than combinations 2, 3 or 5.

The first conclusion we can draw is that the choice of classifier between RT and C4.5 seems to have little effect. When just bagging is employed (hybrid 1 and 4), C4.5 performs much worse than RT. It is not immediately obvious why. For the other two combination pairs (2,5) and (3,6), there is very little difference between RT and C4.5. Just switching Rotation Forest to using RT (i.e. moving from hybrid 6 to 3) actually gives a marginal improvement in average accuracy, although the average rank is lower. This suggests that the base classifier could be changed without significant average loss.

The second conclusion is that the four hybrids that employ some form of PCA (2,3,5,6) are significantly better than the two that do not. The utility of transformation prior to classification is well known. The PCA operation is the core distinctive characteristic of the rotation forest classifier, and these results suggest it is the key component in improving performance. The benefit from the rotation forest approach is that the sparse transformation matrix means it does not need to perform the transformation on the whole data. The switch from bagging with full PCA to rotation forest style PCA in combinations (2,3) and (5,6) does not make a significant difference.

\section{Rotation forest sensitivity analysis}
\label{section:sensitivity}

We stated in Section~\ref{section:comparison} that rotation forest was not sensitive to parameters, given sensible default values, since tuning parameters had no significant effect on accuracy compared to the default values given in Table~\ref{tab:fixed}.
We explore this characteristic in more detail, through an examination of the performance of the algorithm with a range of parameter settings. This has been previously studied by Kuncheva and Rodriguez ~\cite{kuncheva07experimental}, who compared different values for the number of feature subsets and the ensemble size. Their priority was to compare rotation forest against random forest, bagging and adaboost at different parameter settings. Given we tune classifiers, this is of less concern to us. Our interest lies in examining whether it is worth the computational effort of tuning a rotation forest (as opposed to use sensible default values).

\begin{figure}[!ht]
	\centering
       \includegraphics[width =8cm, trim={2.2cm 10cm 0cm 11cm},clip]{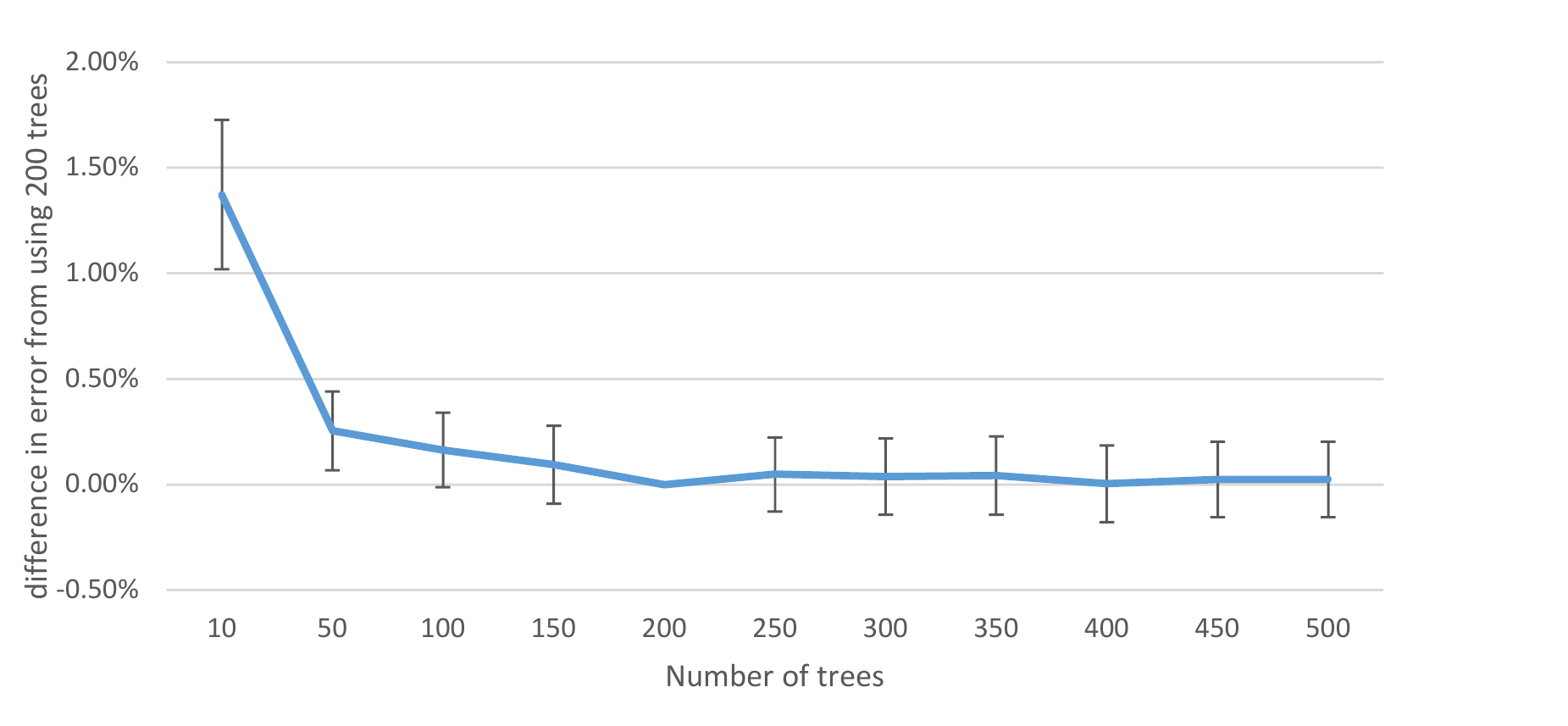}              	
       \caption{The mean difference in error with increasing number of trees when compared to rotation forest with 200 trees. The error bars represent the 95\% confidence intervals for the hypothesis test that the mean difference is zero.}
       \label{fig:numTrees}
\end{figure}

The three parameters we consider are  the number of trees in the ensemble ($k$), the number of features per group ($f$) and the proportion of cases sampled for each group ($p$). Figure~\ref{fig:numTrees} shows the variation in average error on the UCI data when compared to the default value of 200 trees used in the experiments presented in Section~\ref{section:comparison}. Each data point is the average difference over 121 datasets for selecting between 10 and 500 trees. The current default of 10 trees is significantly worse than selecting 50 trees. After 50 trees, the error decreases, but there is no number of trees that is a significant improvement on the default value we have used (200 trees).

\begin{figure}[!ht]
	\centering
       \includegraphics[width =8cm, trim={2.2cm 10cm 0cm 12cm},clip]{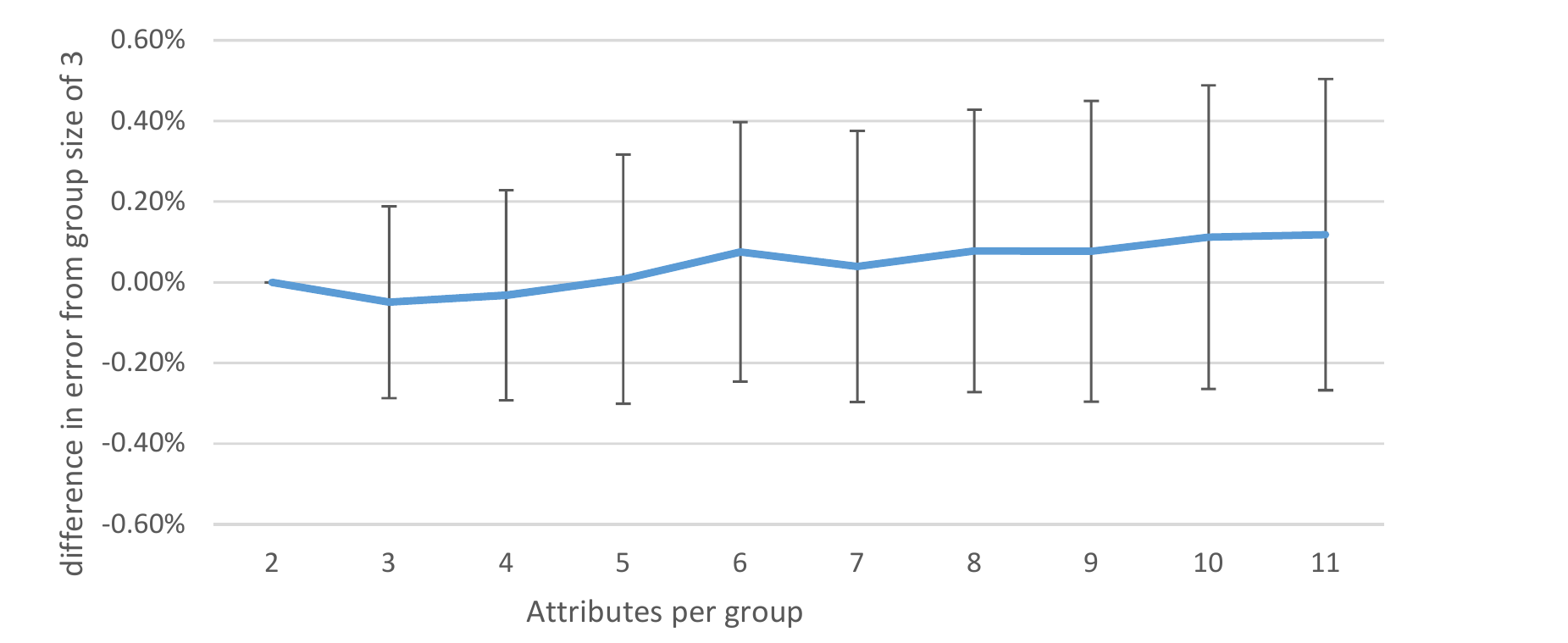}              	
       \caption{The mean difference in error with increasing number of attributes per group compared to rotation forest with 3 per group. The error bars represent the 95\% confidence intervals for the hypothesis test that the mean difference is zero.}
       \label{fig:numGroup}
\end{figure}
\begin{figure}[!ht]
	\centering
       \includegraphics[width =8cm, trim={2.2cm 10cm 0cm 12cm},clip]{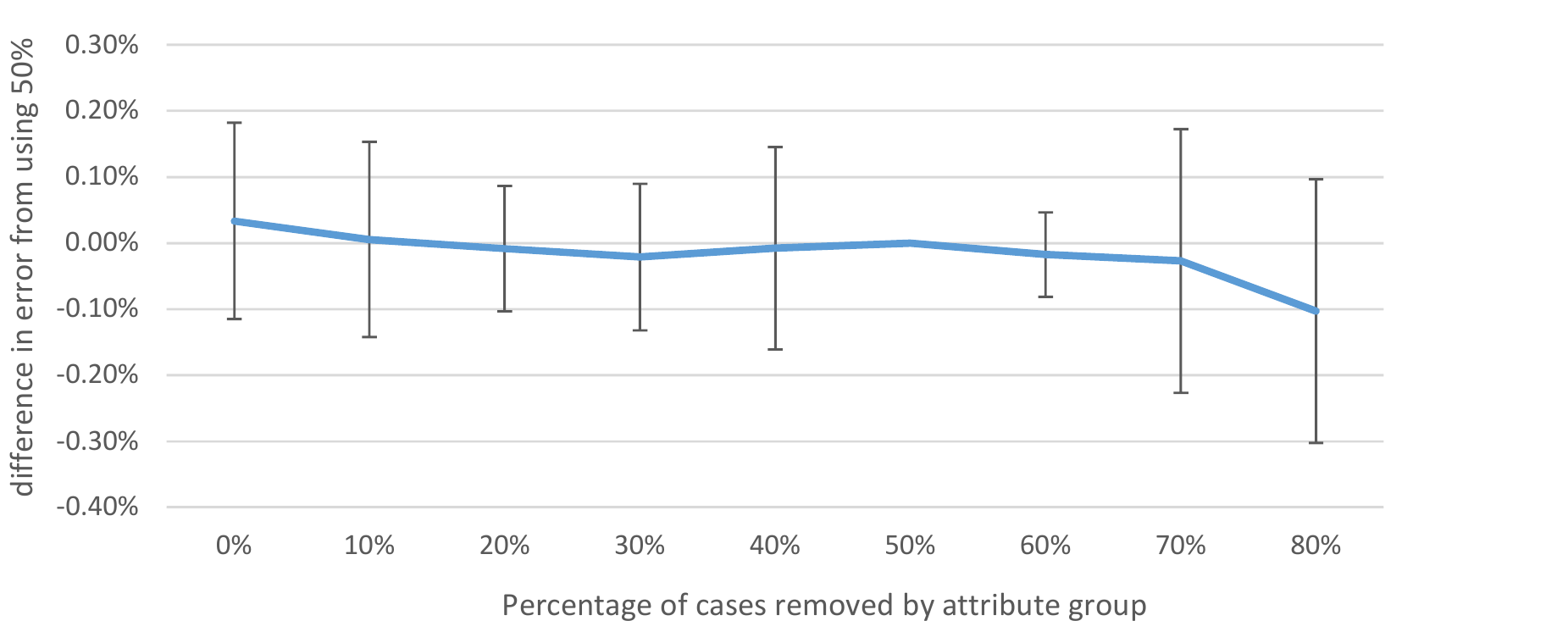}              	
       \caption{The mean difference in error with increasing percentage of cases removed per group, compared to rotation forest with 50\% removed. The error bars represent the 95\% confidence intervals for the hypothesis test that the mean difference is zero.}
       \label{fig:percentageRemoved}
\end{figure}

Figure~\ref{fig:numGroup} shows the effect of selecting an alternative number of attributes to place in each group prior to the application of PCA. The default number of 3 per group gives the lowest error, which gradually increases as more attributes are selected per group, but there is no significant difference between the default and any group size between 4 and 8. Selecting bigger group sizes (9 to 12) increases the error. We think this is due to the small number of attributes in many of the UCI datasets.

Lastly, Figure~\ref{fig:percentageRemoved} shows the change in error as more cases are removed from each group prior to the application of PCA. There is no pattern, and it is perhaps surprising that removing 80\% of cases is not significantly different to removing 50\% or none. We stress the differences in errors in these graphs are very small and that as long as a reasonable number of trees is used in the ensemble, the algorithm is robust to the parameter settings. Hence, we conclude that the default values we use (200 trees, 3 attributes per group, 50\% selected) are as good as any others and rotation forest is robust to variation in these parameters. The Weka default of 10 trees results in significantly higher error, but from 50 to 500 trees there is no significant difference to using 200. Whilst adding more trees may improve performance, that improvement comes at a computational cost. For any given dataset, the question is then how many trees can we afford to use, given the time constraints? This is addressed in Section~\ref{section:timings}.

\section{Contract rotation forest}
\label{section:timings}

\begin{figure}[!ht]
	\centering
       \includegraphics[width =8cm, trim={5cm 4cm 2.5cm 3cm},clip]{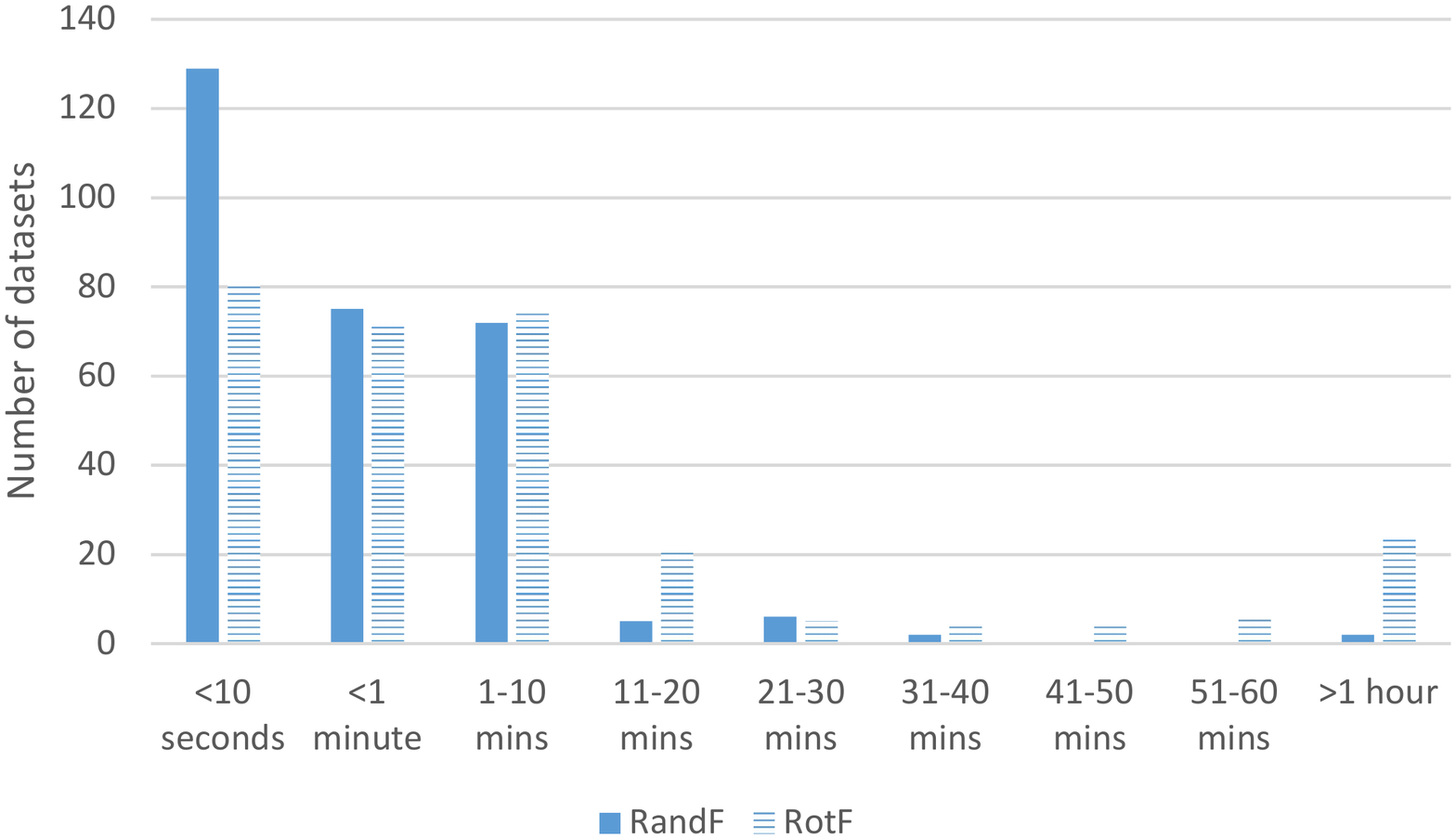}              	
       \caption{Number of problems for given build time ranges for random forest and rotation forest classifiers.}
       \label{fig:buildTime}
\end{figure}

We have demonstrated that rotation forest is on average more accurate than other classifiers on three distinct sets of datasets, and that it does not on average require tuning. At what cost does this extra accuracy come? Specifically, does it take longer on average to build rotation forest than other classifiers? Build times are hard to compare when running experiments on shared architecture using different software.  Furthermore, predicting timings for iterative classifiers is made more difficult because of variation in convergence rates. Because of this, we restrict our attention to comparing build time for random forest and rotation forest. Both of these classifiers were built from the same Weka code base and both were distributed over a computing cluster. To mitigate against variation in build time we take the median time of the 30 resamples of each dataset. It is worth noting that if we do not tune or estimate the error from the train data, none of the problems we have evaluated present too much of a problem. The longest median build time for rotation forest is 33 hours (ST\_ElectricDevices) and for random forest it is 2 hours (miniboone). The median build times over all datasets used in the experiments are summarised in Figure~\ref{fig:buildTime}.

\begin{figure}[htbp]
\centering
    \subfloat[Number of Training Cases ]{\label{fig:relative_a}\includegraphics[trim=2cm 9.7cm 2cm 10.65cm, clip=true,width=0.85\linewidth]{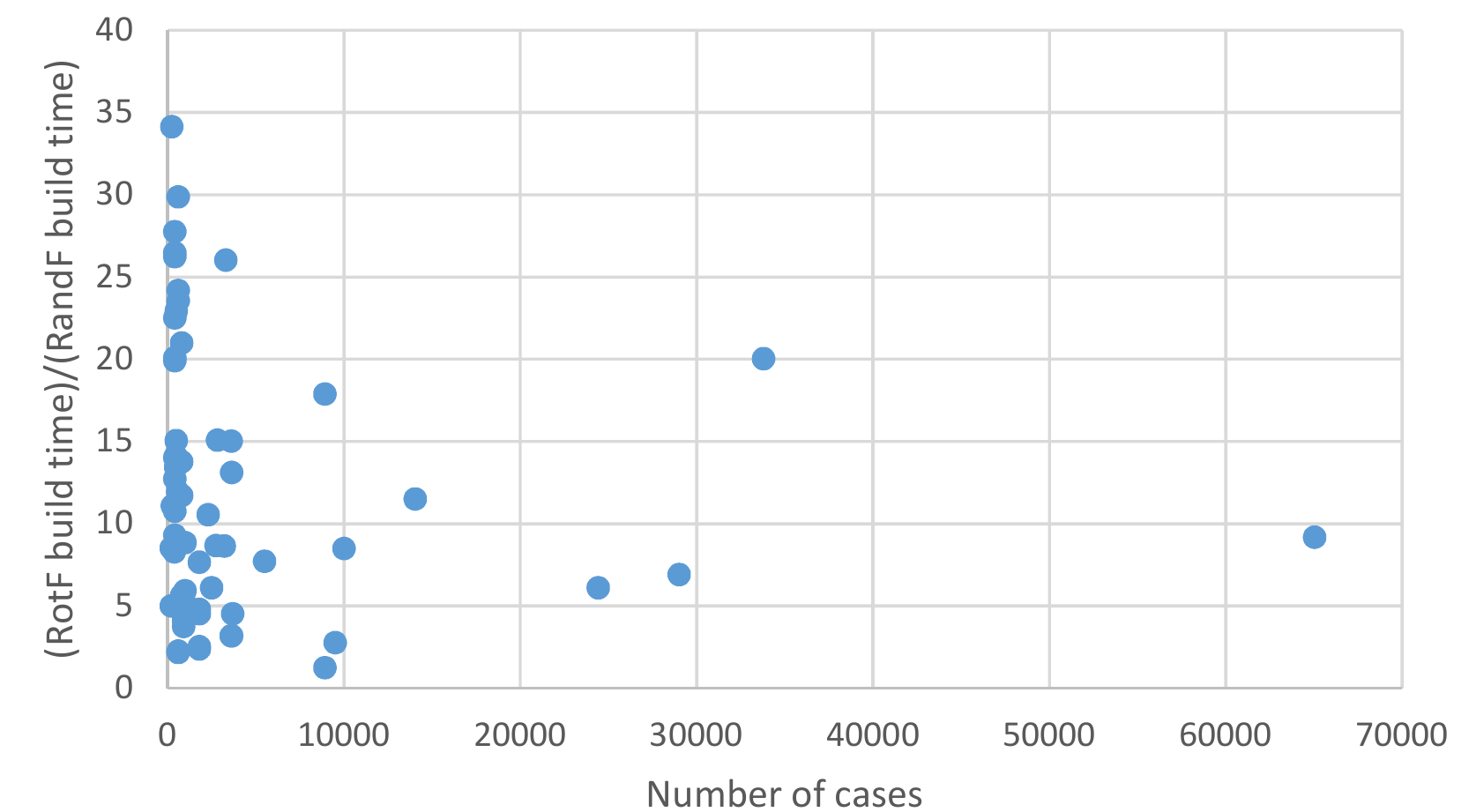}} \\
	\subfloat[Number of Attributes]{\label{fig:relative_b}\includegraphics[trim=2cm 3.8cm 2.2cm 3.4cm, clip=true,width=0.85\linewidth]{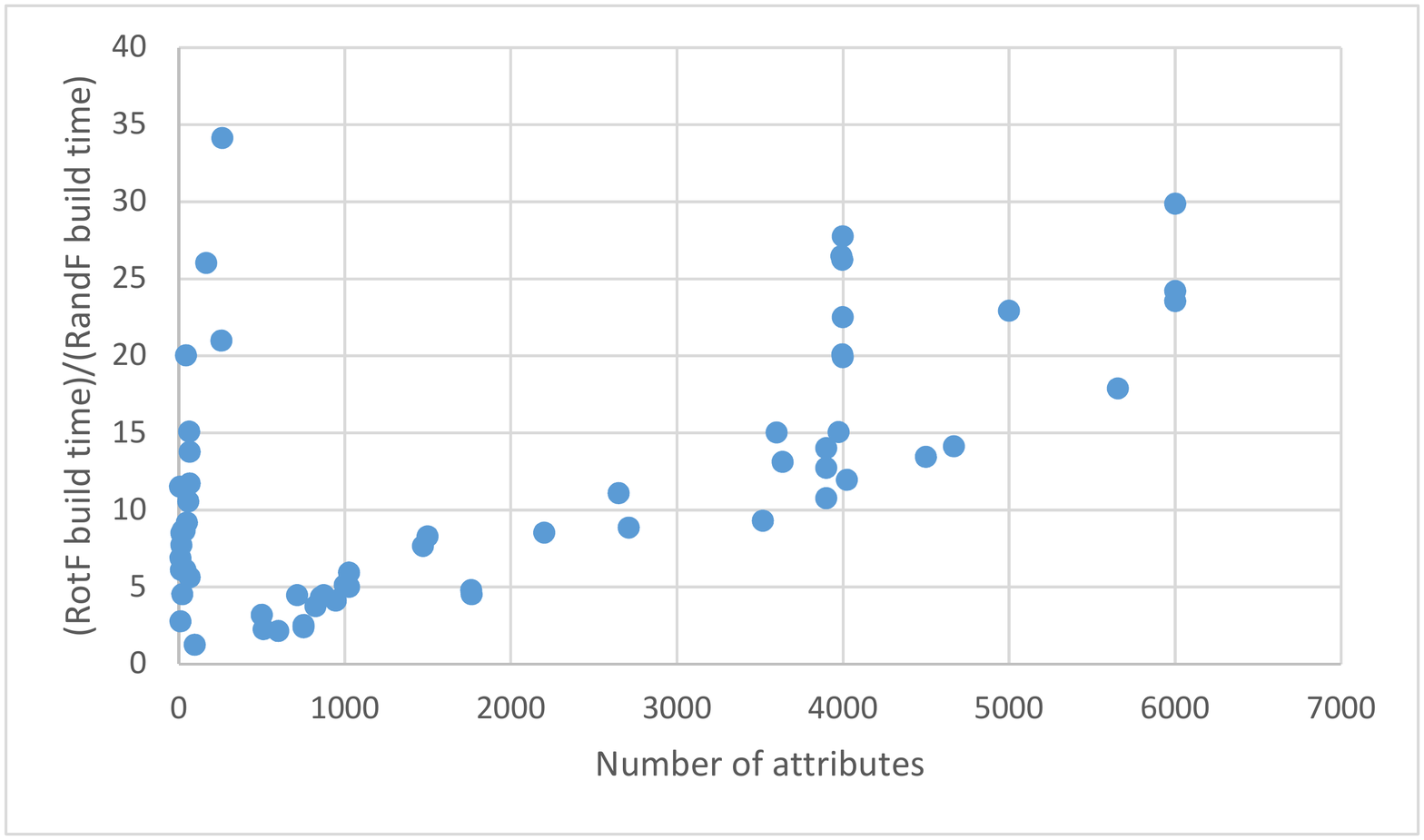}}
	
	\caption{Relative speed of rotation forest and random forest plotted against the number of training cases (a) and the number of attributes (b).}
	\label{fig:relative}
\end{figure}

 The three parameters that most affect build time for the forest classifiers are the number of trees, the number of cases and the number of attributes. Build time will scale linearly with the number of trees. We addressed the influence of using different fixed number of trees on error in Section~\ref{section:sensitivity}. The  key issue in understanding the relative scalability of random forest and rotation forest is understanding how build time changes with number of cases and number of attributes. On average, on problems where rotation forest takes more than 10 minutes, rotation forest with 200 trees is 11.6 times slower than random forest with 500 trees. Figure~\ref{fig:relative} shows the relative speed (rotation forest time divided by random forest time) plotted against the number of cases and the number of attributes.


%

The improvement in accuracy achieved by rotation forest comes at the cost of an order of magnitude more computation. The obvious question is, if we gave random forest the same amount of time, would it become as accurate as rotation forest? Tuning does not help random forest on average, so the easiest way of testing this is to increase the number of trees. We reran the experiments with a  random forest of 10,000 trees. Obviously, a 10,000 tree random forest requires a large amount of memory (three problems required 30 GB), but our focus is on time complexity. In terms of error, rotation forest with 200 trees is still significantly better than random forest with 10,000 trees with approximately equivalent runtime (see spreadsheet RandomForest10000.xls for the results).

Rather than giving random forest more time to build a model, the converse question is, can we restrict the computation time for rotation forest for large problems without loss of accuracy? The most time consuming process of rotation forest is performing PCA on subsets of attributes. This transformation allows the classifier to better capture interactions between attributes, and our ablation experiments demonstrated that this stage of the process is the most important. In contrast to random forest, rotation forest uses all $m$ features for each classifier. Our ablative experiments have shown that reducing the number of features to $\sqrt{m}$ does have a significantly detrimental effect on accuracy. We test whether a less extreme feature space size reduction also results in significant loss in accuracy by the simple expedient of randomly selecting a fixed number of attributes for each tree. Each classifier in the ensemble receives a random subset of the features to train upon. The sampling is exceptionally easy to implement. We simply extend the class \texttt {RotationForest}, and override the method that permutes the attributes prior to grouping so that partitions are formed from a random subset. Thus step 3 in Algorithm~\ref{algo:RotF} is performed with a reduced set of features for each tree. Everything else about the classifiers is identical. We use the same default parameters as before. Our first experiment is designed to test whether sampling attributes makes any difference to accuracy on problems where the number of attributes exceeds the number randomly selected. Of the 121 UCI datasets used in~\cite{delgado14hundreds}, only 23 have more than 40 features. Hence we fix \texttt{maxNumAtts} to 40 for the classifier we denote RotF$_{40}$ and examine whether there is any difference in average accuracy (over 30 resamples with 50\% train and 50\% test) on these data.


The results are shown in table~\ref{tab:randrot1}. The final column provides an estimate of the speed up (time for RotF divided by time for RotF$_{40}$). The highest average accuracy is in bold. There is no significant difference between the mean ranks as tested with a Wilcoxon sign-rank test ($p<0.01$). If a dataset's result is significantly better using a pairwise test on the 30 folds, the result has a star next to it. RotF is significantly better on two problems; RotF$_{40}$ is significantly better on three.
 		\begin{table*}[!htbp]
			\centering
			\caption{Mean (and standard error) for full rotation forest (RotF) compared to rotation forest with a fixed, random number of features selected for each tree (RotF$_{40}$) on 21 UCI datasets. The highest average is in bold. Results with a * are significantly better. }
			\begin{tabular}{l|c|l|l|c} \hline
dataset & \#Attributes & RotF & RotF$_{40}$ & Speed up \\ \hline

arrhythmia & 262	&  {\bf 73.96 ($\pm$0.34)*} 	& 68.14 ($\pm$0.26)  	&  7.7	\\
semeion & 256	& 90.73 ($\pm$0.21)  	& {\bf 93.17 ($\pm$0.16)* } 	&  6.21	\\
musk-1 & 166	& 88.3 ($\pm$0.42)  	& {\bf 89.21 ($\pm$0.37) } 	&  4.29	\\
musk-2 & 166	& {\bf 97.46 ($\pm$0.05)* } 	& 96.9 ($\pm$0.07)  	&  4.75	\\
hill-valley & 100	& {\bf 67.99 ($\pm$0.35) } 	& 66.85 ($\pm$0.36)  	&  2.93	\\
low-res-spect & 100	& 89.19 ($\pm$0.32)  	& {\bf 89.95 ($\pm$0.25) } 	&  2.5	\\
libras & 90	& 82.7 ($\pm$0.54)  	& {\bf 83.61 ($\pm$0.53) } 	&  2.15	\\
ozone & 72	& {\bf 97.05 ($\pm$0.02) } 	& {\bf 97.05 ($\pm$0.02) } 	&  1.76	\\
plant-margin & 64	& 83.18 ($\pm$0.18)  	& {\bf 83.19 ($\pm$0.22) } 	&  1.55	\\
plant-shape & 64	& 66.72 ($\pm$0.22)  	& {\bf 66.96 ($\pm$0.28) } 	&  1.45	\\
plant-texture & 64	& 80.56 ($\pm$0.2)  	& {\bf 81.38 ($\pm$0.19)*} 	&  1.58	\\
optical & 62	& 98.04 ($\pm$0.04)  	& {\bf 98.17 ($\pm$0.05) } 	&  1.45	\\
conn-bench-sonar & 60	& {\bf 82.57 ($\pm$0.62) } 	& 82.16 ($\pm$0.62)  	&  1.53	\\
molec-biol-splice & 60	& 94.17 ($\pm$0.1)  	& {\bf 94.7 ($\pm$0.08) *} 	&  1.28	\\
synthetic-control & 60	& 98.3 ($\pm$0.14)  	& {\bf 98.59 ($\pm$0.11) } 	&  1.42	\\
audiology-std & 59	& {\bf 79.83 ($\pm$0.6) } 	& 79.77 ($\pm$0.66)  	&  1.43	\\
molec-biol-promoter & 57	& 84.63 ($\pm$0.94)  	& {\bf 84.88 ($\pm$0.87) } 	&  1.51	\\
spambase & 57	& 94.85 ($\pm$0.08)  	& {\bf 95.06 ($\pm$0.06) } 	&  1.4	\\
lung-cancer & 56	& {\bf 50.59 ($\pm$2.01) } 	& 49.61 ($\pm$1.91)  	&  1.19	\\
spectf & 44	& {\bf 87.94 ($\pm$0.35) } 	& 87.36 ($\pm$0.5)  	&  1.09	\\
oocytes-merluccius & 41	& 83.5 ($\pm$0.21)  	& {\bf 83.86 ($\pm$0.26) } 	&  1.02	\\ \hline
wins & 		 & 7 & 13 & 	\\ \hline
\end{tabular}
			\label{tab:randrot1}
		\end{table*}
Table~\ref{tab:randrot1} suggests there is no overall detrimental effect from sampling the features and significant speed up can be achieved. However, these datasets are all relatively small. To test on data with larger feature spaces, we repeat the experiment on the UCR datasets. If we randomly select 100 features (RotF$_{100}$) rather than use them all for the 21 problems with the largest number of features, there is no significant overall difference in accuracy on these data (see Figure~\ref{tab:randrot2}). RotF wins on 16, RotF$_{100}$ on 11 and there is one tie.  and yet we get a speedup of between 7 and 61 times. In a pairwise comparison, RotF is significantly better on 2 problems, RotF$_{100}$ on 1.

		\begin{table*}[!htbp]
			\centering
			\caption{Mean (and standard error) for full rotation forest (RotF) compared to rotation forest with a fixed, random number of features selected for each tree (RotF$_{100}$) on 21 UCR datasets.}
			\begin{tabular}{l|c|l|l|c} \hline
dataset & \#Attributes & RotF & RotF$_{100}$ & Speed up \\ \hline
HandOutlines & 2709	& {\bf 91.04 ($\pm$0.25) } 	& 90.57 ($\pm$0.28)  	&  60.94	\\
InlineSkate & 1882	& 34.84 ($\pm$0.34)  	& {\bf 35.73 ($\pm$0.4) } 	&  30.34	\\
CinCECGtorso & 1639	& 70.58 ($\pm$1.11)  	& {\bf 78.81 ($\pm$1.11) *} 	&  31.59	\\
Haptics & 1092	& {\bf 47.66 ($\pm$0.4) } 	& 46.65 ($\pm$0.31)  	&  16.42	\\
Mallat & 1024	& {\bf 95.15 ($\pm$0.32) } 	& 94.67 ($\pm$0.42)  	&  9.44	\\
Phoneme & 1024	& 14.2 ($\pm$0.15)  	& {\bf 14.7 ($\pm$0.15) } 	&  13.9	\\
StarlightCurves & 1024	& {\bf 97.01 ($\pm$0.04) } 	& 96.85 ($\pm$0.04)  	&  15.64	\\
UWaveAll	& 945	& 94.83 ($\pm$0.07)  	& {\bf 95.12 ($\pm$0.07) } 	&  11.89	\\
Worms & 900	& {\bf 60.17 ($\pm$0.8) } 	& 60.09 ($\pm$0.78)  	&  13.51	\\
WormsTwoClass & 900	& {\bf 65.84 ($\pm$0.86) } 	& 64.89 ($\pm$0.78)  	&  14.16	\\
NonInvasive1 & 750	& {\bf 90.6 ($\pm$0.09)*} 	& 89.41 ($\pm$0.1)  	&  7.64	\\
NonInvasive2 & 750	& {\bf 93.27 ($\pm$0.09)*} 	& 92.37 ($\pm$0.07)  	&  7.58	\\
Computers & 720	& 67.32 ($\pm$0.37)  	& {\bf 67.44 ($\pm$0.51) } 	&  10.31	\\
LargeKitchen & 720	& 62.57 ($\pm$0.34)  	& {\bf 63.38 ($\pm$0.31) } 	&  9.61	\\
RefrigerationDevices & 720	& 58.06 ($\pm$0.4)  	& {\bf 58.28 ($\pm$0.42) } 	&  9.73	\\
ScreenType & 720	& {\bf 46.87 ($\pm$0.35) } 	& 46.08 ($\pm$0.36)  	&  9.84	\\
SmallKitchen & 720	& {\bf 72.22 ($\pm$0.38) } 	& 71.26 ($\pm$0.34)  	&  9.85	\\
Lightning2 & 637	& 76.07 ($\pm$0.96)  	& {\bf 77.21 ($\pm$0.73) } 	&  8.79	\\
Car & 577	& 77.33 ($\pm$1.01)  	& {\bf 78.89 ($\pm$0.86) } 	&  8	\\
OliveOil & 570	& {\bf 90.56 ($\pm$0.82) } 	& 89.67 ($\pm$0.89)  	&  7.33	\\
BeetleFly & 512	& 80 ($\pm$1.44)  	& {\bf 80.33 ($\pm$1.55) } 	&  8.37	\\
\hline
wins & 		 & 11 & 10 & 	\\ \hline

\end{tabular}
			\label{tab:randrot2}
		\end{table*}

Our results indicate that randomly selecting attributes does not on average decrease accuracy, but for when a very small proportion of the total number of attributes is selected ($\sqrt{m}$), it is detrimental. The question is, to what degree should we sample the attributes for each tree? This is obviously dependent on the amount of computational time we can allocate to building the classifier. We desire a contract classifier, where we set the level of attribute sampling required for a given problem in order to allow rotation forest to run in approximately given amount of time. To answer this question, we need to estimate how long rotation forest will take for a given dataset, then estimate the speed up we will get for a given number of attributes sampled. We can then calculate the required speed up and estimate the number of features that will provide this speed up.

The runtime complexity of building a rotation forest is governed by: the data characteristics (number of attributes, $m$, the number of cases, $n$, and the number of classes, $c$); the rotation forest parameters (number of trees, $t$, number of feature sets, $r$, the number of features per set $f$, and the proportion selected for each set, $p$); and the complexity of building the resulting C4.5 decision tree. Decision trees are data dependent heuristics, so it is hard to estimate the complexity. Given we are using real-valued attributes and C4.5 uses an information gain splitting criteria, the sorting of attributes is likely to dominate the complexity. The best case for C4.5 is when the root node is the only node built, and the complexity is $O(mnlog(n))$. The worst case for C4.5 would be if the final tree has the minimum number of cases at each node (default value of 2). In this case the tree will have depth $O(n)$, and the worst case of a basic implementation is $O(mn^2log(n))$. However, assuming the sensible use of indexing minimizes the need for resorting, the work done on the root node is likely to dominate the run time. For rotation forest, C4.5 is built with $rf=m$ attributes and a proportion of $n$ cases, so we characterise the run time as

\begin{equation}
t_{c45}(n,m) = \beta_1 \cdot m + \beta_2 \cdot n+ \beta_3 \cdot m \cdot n \cdot log(n)+\epsilon,
\label{eq:c45}
\end{equation}
where $\beta_1$ is an unknown constant dependent on $p$, and $\epsilon$ is a independent random variable assumed to be normally distributed. Before rotation forest applies a decision tree, it transforms the dataset. The core PCA operation that is applied to each of the $r$ sets is complexity $O(ab^2+b^3)$, for $a$ cases and $b$ features. Rotation forest has $f$ features per set, and each set contains a different number of cases based on the resampling proportion $p$ and the number of classes selected for that set. We assume the class selection probability and the proportion of cases sampled are fixed to $p=0.5$.  Hence, the expected run time complexity for a single set is $O(nf^2+f^3)$, which, assuming $n$ is large and $f$ is small (f defaults to 3), will be dominated by the term $O(nf^2)$.
For the full rotation forest, $rf=m$.

$$ t_{rf}(n,m)=t_{c45}(n,m,p) + \beta_4 \cdot r\cdot n \cdot p \cdot f^2 + \epsilon.$$

We assume $f$ and $p$ are constants, and that there is a constant amount of work to do for all problems, so we can simplify to

$$ t_{rf}(n,m,r,f)=\beta_0+  t_{c45}(n,m,p)+ \beta_4 \cdot m \cdot n +\epsilon.$$
Substituting Equation~\ref{eq:c45} gives us
$$ t_{rf}(n,m)=\beta_0+ \beta_1 \cdot m + \beta_2 \cdot n+ \beta_3 \cdot m \cdot n \cdot log(n)+ \beta_4 \cdot m \cdot n + \epsilon.$$

If we assume  a  normal distribution we can estimate the parameters from experimental data using using linear regression. The model is a coarse approximation, but we only require an approximate model that can give an indication of whether it is possible to fit a rotation forest on a given dataset. We extract the timing results for all problems that took more than 30 minutes for a single run and fit the following linear regression model

\begin{equation} t_{rf}(n,m)=0.64+\frac{0.132}{1000}\cdot n+\frac{0.246}{1000}\cdot m+\frac{0.615}{1,000,000}\cdot {m\cdot n}.
\label{eq:timingModel}
\end{equation}

We have dropped the $ m \cdot n \cdot log(n)$ term for simplicity. The model has an adjusted $R^2$ of 96\% and there is no obvious pattern in the residuals. It is not particularly accurate, with a mean absolute error of approximately one hour. However, the predicted value is within the range of observed time for all datasets, and error in the order of magnitude of hours is acceptable for our requirements. We use the timing model given in Equation~\ref{eq:timingModel} to form a 95\% confident prediction interval for the run time, using the standard formula derived from the covariance matrix $(X^TX)^{-1}$,

\begin{equation}
\hat{y}_0-s\cdot t\sqrt{1+x^T_0 (X^TX)^{-1}x_0}<y_0<\hat{y}_0+s\cdot t\sqrt{1+x^T_0 (X^TX)^{-1}x_0}
\label{eq:predictionInterval}
\end{equation}
where $x_0$ are the observed dependent variables for the new instance, $\hat{y}_0$ is the predicted time from model~\ref{eq:timingModel}, $X$ is training matrix of regressor values, $s$ the standard error of the trained model and $t$ is the value of the t distribution corresponding to $\alpha=0.05$. Timing is of course machine dependent. To scale from one machine to another, we use a reference operation to calibrate the timing model to give a scaling factor.


To make rotation forest more scalable, we adapt the algorithm to handle cases where the training set is very large in terms of number of cases and/or number of features. We use a simple heuristic to constrain the algorithm to train within a contracted time limit which utilises the timing model. Algorithm~\ref{algo:ContractRotF} describes our contract version of rotation forest. In line 1 we estimate the 95\% upper prediction interval using Equation~\ref{eq:predictionInterval}. If we predict a build time less than the contract time $t$, we can simply build rotation forest normally. If not, we make the practical decision to reduce features if the number of features is greater than the number of training cases, and vice versa otherwise. We require a minimum number of trees in the ensemble, $e_{min}$, which we default to 50 based on the sensitivity analysis presented in Figure~\ref{fig:numTrees} and a maximum, $e_{max}$, which we set to 200. We estimate the maximum number of attributes we can have to build $e_{min}$ trees on line 5. To compensate for variability in the timing model, we select between $e_{min}/2$ and $e_{min}$ attributes for each of the first $e_{min}$ trees (line 8). We then build a limited attribute tree in the manner described above. Because there is large variability in observed build times, we adapt the predicted build time based on the observed build time using a simple form of reinforcement learning (line 13). This approach is simplistic, but a reasonable first approximation.

\begin{algorithm}[!ht]
	\caption{contractRotationForest(Data $D$, time limit $t$)}
\label{algo:ContractRotF}
	\begin{algorithmic}[1]
\Require number of attributes ($m$), number of cases ($n$), minimum ensemble size ($e_{min}$), maximum ensemble size ($e_{max}$), learning rate ($\alpha$).
\State Let ${\bf F}$ be the set of C4.5 trees, initially empty.
\State $\hat{t} \leftarrow$ estimateTimeUpperBound($D$)
\If {$\hat{t}<t$}
\State buildRotationForest($D$)
\Return
\EndIf
\If{$m<n$}
\State $\hat{m} \leftarrow$ estimateMaxNosAttributes($m,e_{min},\hat{t},t$)
\State $s \leftarrow 0, e \leftarrow 0$
\While{$s < t \wedge e < e_{min}$ }
\State $k \in [\hat{m}/2 \ldots \hat{m}]$
\State ${\bf F} \leftarrow {\bf F}\; \cup $  buildRandomAttributeRotationTree($D,k$)
\State $b \leftarrow$ getTime()
\State $s \leftarrow$ updateTotalBuildTime($b$)
\State $\hat{t} \leftarrow (1-\alpha) \cdot \hat{t}+\alpha \cdot b $
\State $\hat{m} \leftarrow$ estimateMaxNosAttributes($m,e_{min},\hat{t},t$)
\EndWhile
\While{$s < t \wedge e < e_{max}$ }
\State $k \in [\hat{m} \ldots m]$
\State ${\bf F} \leftarrow {\bf F}\; \cup $  buildRandomAttributeRotationTree($D,k$)
\State $s \leftarrow$ updateBuildTime()
\EndWhile
\Else
\State $\hat{n} \leftarrow$ estimateMaxNosCases($n,e_{min},\hat{t},t$)
\State $s \leftarrow 0, e \leftarrow 0$
\While{$s < t \wedge e < e_{min}$ }
\State $k \in [\hat{n}/2 \ldots \hat{n}]$
\State ${\bf F} \leftarrow {\bf F}\; \cup $ buildRandomCaseRotationTree($D,k$)
\State $s \leftarrow$ updateBuildTime()
\EndWhile
\While{$s < t \wedge e < e_{max}$ }
\State $k \in [\hat{n} \ldots n]$
\State ${\bf F} \leftarrow {\bf F}\; \cup $ buildRandomCaseRotationTree($D,k$)
\State $s \leftarrow$ updateBuildTime()
\EndWhile

\EndIf
\end{algorithmic}
\end{algorithm}

To test the model, we use a set of datasets not used at any stage of experimentation so far. These large datasets are part of a new multivariate time series classification archive~\cite{bagnall18mtsc}. Ultimately, these data will be used for testing bespoke algorithms for multivariate time series classification. However, as a basic sanity check for benchmarking purposes, we wish to evaluate standard classifiers on this data first. The most naive approach for multivariate time series data is to concatenate all the dimensions to make univariate time series problems, then use a standard classifier. Given the good performance of rotation forest on time series classification problems, it seems reasonable to benchmark with rotation forest. Concatenated time series problems often have very large feature spaces.  Tables~\ref{tab:small} and ~\ref{tab:large} summarise the data with which we test the contract rotation forest. Rotation forest can be built in under 12 hours on the smaller problems described in Table~\ref{tab:small}, whereas the classifier takes more than 12 hours to build on the larger problems summarised in Table~\ref{tab:large}.

		\begin{table}[!htbp]
			\centering
			\caption{Problems where full rotation forest completes in under 12 hours}
			\label{tab:small}
			\begin{tabular}{l|l|ccc} \hline
ID& Dataset                 & \# Attributes & \#  Cases & \# Classes \\ \hline
1 & Cricket                 &    7,182       &    108    &      12     \\
2 & EthanolConcentration    &    5253        &    261    &      4     \\
3 & HandMovementDirection   &    4000        &    320    &      4     \\
4 & SelfRegulationSCP1      &    5376        &    268    &      2     \\
5 & SelfRegulationSCP2      &    5376        &    268    &      2     \\
6 & StandWalkJump           &    10,000      &    12     &      3     \\
\hline
\end{tabular}
\end{table}

		\begin{table}[!htbp]
			\centering
			\caption{Problems where full rotation forest completes in over 12 hours}
			\label{tab:large}
			\begin{tabular}{l|l|ccc} \hline
 ID & Dataset                 & \# Attributes & \# Cases & \# Classes \\ \hline
 7 & EigenWorms              &    107,904     &    131    &      5     \\
 8 & FaceDetection           &    8928        &    5890   &      2     \\
 9 & Heartbeat               &    24,705      &    204    &      2     \\
 10 & MotorImagery            &    192,000     &    278    &      2     \\
11 & PEMS-SF                 &    138,672     &    267    &      7     \\
 \hline
\end{tabular}
\end{table}


There are two features of contract rotation forest we wish to test with this data: does the classifier complete in approximately the contracted time, and what happens to the accuracy with increased build time? Tables~\ref{tab:contractsmall} and~\ref{tab:contractlarge} show the average build time against contracted build time.

\begin{table}[!htbp]
\centering
\caption{Observed build time for increasing contract time for problems rotation forest completes in under 12 hours. Names for given problem numbers are in Table~\ref{tab:small}}
\label{tab:contractsmall}
\begin{tabular}{c|cccccc} \hline
Contract Time (mins) & 1	 & 2	 & 3	 & 4	 & 5	 & 6	 	\\ \hline
5	 & 4.85	     & 4.64	 & 4.65	 & 4.6	 & 4.65	 & 3.62	 \\
30	 & 24.58	 & 27.35	 & 27.31	 & 27.28	 & 27.32	 & 3.16	 \\
60	 & 20.78	 & 41.09	 & 44.61	 & 39.45	 & 54.58	 & 2.95	 \\
120	 & 22.96	 & 47.86	 & 49.68	 & 40.21	 & 60.09	 & 3.21	 \\
180	 & 24.38	  & 46.87	 & 49.88	 & 42.5	 & 71.39	 & 3.28	 \\
720	 & 25.25	  & 45.47	 & 58.66	 & 42.25	 & 65.47	 & 3.79	 \\
1440 & 32.85	  & 49.76	 & 48.13	 & 40.97	 & 48	 & 4.26	 \\\hline               			
\end{tabular}
\end{table}

\begin{table}[!htbp]
\centering
\caption{Observed build time for increasing contract time for problems where rotation forest takes more than a 12 hours to complete. Names for given problem numbers are in Table~\ref{tab:large}}
\label{tab:contractlarge}
\begin{tabular}{c|ccccc} \hline
Contract Time (mins) & 7	 & 8	 & 9	 & 10	 & 11	\\  \hline
5	 & 5.08	 & 2.71	 & 5.11	 & 5.16	 & 4.82	 \\
30	 & 28.4	 & 28.42	 & 28.3	 & 28.61	 & 28.04	 \\
60	 & 56.05	 & 41.77	 & 56	 & 56.4	 & 56.08	 \\
120	 & 111.25	 & 81.29	 & 109.43	 & 112.12	 & 112.63	 \\
180	 & 169.36	 & 121.54	 & 159.42	 & 167.94	 & 169.02	 \\
720	 & 588.88	 & 480.96	 & 	701.33        & 568.77	 & 572.7	 \\
1440 & 	1210.58  & 965.89	 & 	1321.99        & 560.27	 & 1029.46	 \\	 \hline               			
\end{tabular}
\end{table}
The results demonstrate that the contract is being enforced within an acceptable tolerance. Smaller problems that can complete within the contract take approximately the same amount of time to run, independent of the contract. Problems unable to complete within the contract are completing within the contract time.

\begin{table*}[!htbp]
\centering
\caption{Mean accuracy for increasing contract time for problems rotation forest completes in under 12 hours. Names for given problem numbers are in Table~\ref{tab:small}. The shortest contract time where a full model of 200 trees is built is shown in bold. }
\label{tab:accSmall}
\begin{tabular}{c|cccccc} \hline
Contract Time (mins) & 1	 & 2	 & 3	 & 4	    & 5	          & 6	 	\\  \hline
5	  & 93.48\%	        & 61.86\%	    & 70.2\%	    & 86.66\%	  & 48.89\%	  & 32.5\%	  \\
30	  & 94.27\%	        & 65.25\%	    & 71.29\%	    & 87.06\%	  & 50.39\%	  & 34.67\%	  \\
60	  & {\bf 93.61\%}	& {\bf 66.73\%}	  & 71.16\%	    & 87.13\%	  & 49.67\%	  & 37.33\%	  \\
120	  & 94.34\%		    & 66.88\%	    & {\bf 72.04\%}	& {\bf87.13\%}& {\bf 50.83\%}	  & {\bf 46.67\%}	  \\
180	  & 93.65\%		    & 66.92\%	    & 72.72\%	    & 87.27\%	  & 50.06\%	  & 43.33\%	  \\
720	  & 93.65\%		    & 66.92\%	    & 72.72\%	    & 87.27\%	  & 50.06\%	  & 43.33\%	  \\
1440	  & 93.61\%	    & 66.92\%	    & 72.72\%	    & 87.27\%	  & 50\%	  & 38\%	  \\
RandF	  & 94.03\%	    & 43.8\%	    & 70.27\%	    & 84.2\%	  & 49.17\%	  & 36.67\%	  \\
RotF	  & 93.06\%	   & 66.92\%	    & 71.02\%	    & 87.2\%	  & 48.89\%	  & 33.33\%	  \\
C4.5	  & 71.25\%	   & 38.78\%	  & 64.22\%	  & 77.68\%	  & 46.5\%	  & 32\%	  \\

\hline               			
\end{tabular}
\end{table*}

\begin{table}[!htbp]
\centering
\caption{Mean accuracy for increasing contract time for problems rotation forest completes in over one day. Names for given problem numbers are in Table~\ref{tab:large}.}
\label{tab:accLarge}
\begin{tabular}{c|ccccc} \hline
Contract Time (mins) & 7	 & 8	 & 9	 & 10	 & 11	\\  \hline
5	  & 62.34\%	  & 58.17\%	  & 69.06\%	  & 49.97\%	  & 93.53\%	  \\
30	  & 59.39\%	  & 61.38\%	  & 72.39\%	  & 47.6\%	  & 94.91\%	  \\
60	  & 59.69\%	  & 60.85\%	  & 73.02\%	  & 50.1\%	  & 95.55\%	  \\
120	  & 61.07\%	  & 60.9\%	  & 73.12\%	  & 48.4\%	  & 95.55\%	  \\
180	  & 59.92\%	  & 61.76\%	  & 75.02\%	  & 49.4\%	  & 97.11\%	  \\
720	  & 58.97\%	  & 63.29\%	  & 76.13\%	  & 48.3\%	  & 96.24\%	  \\
1440	  & 59.77\%	  & 64.47\%	  & 74.32\%	  & 47.5\%	  & 97.4\%	  \\
\hline               			
\end{tabular}
\end{table}

Tables~\ref{tab:accSmall} and~\ref{tab:accLarge} show the accuracy of the classifier with changing contract time. We stress that improved accuracy is not our goal: our intention is to improve usability. However, we can make some qualitative comments on these results. On the small problems (Table~\ref{tab:accSmall}), contracting has little effect. The full model builds in around 1-2 hours for all of these problems. Although tuning has no benefit for rotation forest on average, but it may be desirable on a specific problem. If a single build takes 2 hours, tuning over a large range of parameters could take weeks. These results indicate that we could probably perform the grid search for tuning using a low contract time for each parameter setting. This could reduce the tuning time from days to hours and allow for a wider search space. The results shown in Table~\ref{tab:accLarge} demonstrate two things. Firstly, for problems where the approach of concatenation may have some value (problems 8 and 11), the accuracy of contract rotation forest increases with contract time. For problems where time domain classifiers are inappropriate (7 and 10), the accuracy changes very little. In summary, the contract rotation forest improves the usability of rotation forest in a way that allows for the application of the algorithm to domains where the full model  will not build in reasonable time. Furthermore, It facilitates an exploratory analysis that may give insights into whether the data representation is suitable.

We have not addressed the memory constraints of rotation forest. Rotation forest is memory intensive, particularly when all the attributes are used. We have mitigated against this for contract rotation forest by stopping the build when memory requirements hits a user defined limit (set to 10GB in all experiments). This simple expedient means that the build completes, although it may be better to more intelligently use the memory constraints in the contract process.

\section{Conclusions}
\label{section:conclusions}
Rotation forest is less well known and less frequently used than algorithms such as SVM, random forest and neural networks. Our primary contribution is to demonstrate that, for problems with real-valued attributes, rotation forest should at least be considered as a starting point solution. It is significantly more accurate on average than the alternatives on all the data we have used in the evaluation. We are not suggesting that rotation forest makes other algorithms redundant. The average differences in error are not huge and there is variability over problems. Nevertheless, we believe our experimental results mean that rotation forest should be reevaluated by machine learning practitioners who may not have been aware of its existence or known of how well it performs. To help facilitate the more widespread use of this algorithm we have provided a basic scikit implementation. The main drawback with rotation forest is that it is relatively slow to build, particularly when the data has a large number of attributes. To address this problem we have developed a Weka-based contracted rotation forest classifier. This contract mechanism for rotation forest could be further improved with a more sophisticated reinforcement learning mechanism and serialised checkpointing to allow for continued building after termination. The contract classifier is particularly useful for problems with large attribute sizes, but we believe that refinements to both the contract mechanism and the basic structure of rotation forest may yield further improvements in terms of accuracy, speed and memory usage. Our ablative study suggests a bagged version may not reduce accuracy. This would allow for fast estimation of test set error through out of bag error rather than cross-validation. We have only looked at real-valued problems because rotation forest is based on a real-valued transformation, but a performance evaluation on problems with discrete attributes would also be of interest.

\section*{Acknowledgements}
This work is supported by the UK Engineering and Physical Sciences Research Council (EPSRC)  [grant number EP/M015807/1] and the Biotechnology and Biological Sciences Research Council [grant number BB/M011216/1]. The experiments were carried out on the High Performance Computing Cluster supported by the Research and Specialist Computing Support service at the University of East Anglia and using a Titan X Pascal donated by the NVIDIA Corporation.

\section*{Appendix A: dataset details}

\renewcommand{\arraystretch}{0.88} 
\begin{table}[htb]
\scriptsize
	\centering
	\caption{A full list of the 39 UCI datasets used in the experiments in Section~\ref{section:comparison}. A single zip of these data are available on the accompanying website. }
	\label{table:uciDatasets}
    \setlength\tabcolsep{3pt} 
	\begin{tabular}{|r|c|c|c|}
		\hline
Dataset	&	\#Attributes	&	\#Classes	&	\#Cases \\ \hline
bank	&	16	&	2	&	4521	\\	
blood	&	4	&	2	&	748	    \\	
breast-cancer-w-diag	&	30	&	2	&	569 \\
breast-tissue	&	9	&	6	&	106	\\
cardio-10clases	&	21	&	10	&	2126\\	
conn-bench-sonar-mines-rocks & 60 & 2 & 208 \\
conn-bench-vowel-deterding  & 11 & 11 & 990  \\
ecoli	&	7	&	8	&	336	\\
glass	&	9	&	6	&	214	 \\
hill-valley	&	100	&	2	&	1212	\\
image-segmentation	&	18	&	7	&	2310	\\
ionosphere	&	33	&	2	&	351	\\
iris	&	4	&	3	&	150		 \\
libras	&	90	&	15	&	360	\\
magic   &  10   & 2 & 19020 \\
miniboone & 50 & 2 & 130064\\
oocytes\_m\_nucleus\_4d	&	41	&	2	&	1022 \\		
oocytes\_t\_states\_5b	&	32	&	3	&	912 \\
optical	&	62	&	10	&	5620 \\
ozone	&	72	&	2	&	2536 \\
page-blocks	&	10	&	5	&	5473 \\
parkinsons	&	22	&	2	&	195 \\
pendigits	&	16	&	10	&	10992 \\
planning	&	12	&	2	&	182 \\
post-operative	&	8	&	3	&	90 \\
ringnorm	&	20	&	2	&	7400 \\
seeds	&	7	&	3	&	210 \\
spambase	&	57	&	2	&	4601 \\	
statlog-landsat	&	36	&	6	&	6435 \\
statlog-shuttle	&	9	&	7	&	58000 \\
statlog-vehicle & 18     & 4 & 846 \\
steel-plates	&	27	&	7	&	1941 \\
synthetic-control	&	60	&	6	&	600 \\
twonorm	&	20	&	2	&	7400 \\
vertebral-column-3clases	&	6	&	3	&	310 \\
wall-following	&	24	&	4	&	5456 \\
waveform-noise & 40 & 3 & 5000 \\
wine-quality-white	&	11	&	7	&	4898 \\
yeast	&	8	&	10	&	1484 \\
		
		\hline
	\end{tabular}
\end{table}
\renewcommand{\arraystretch}{1}
\begin{table}[htb]
	\centering
    \scriptsize
	\caption{The 85 UCR time series classification problems used in the experiments for Section~\ref{section:ucr}. A single zip of these data are available on the accompanying website. }
	\label{table:TSCproblems}
    \setlength\tabcolsep{1pt} 
	\begin{tabular}{|r|c|c|c|c|r|c|c|c|c|}
		\hline
		Dataset	&	Atts	&	Classes	&	Train	& Test	&	Dataset	&	Atts	&	Classes	&	Train	&Test \\ \hline
		Adiac	&	176	&	37	&	390	&	391	&	MedicalImages	&	99	&	10	&	381	&	760 \\
		ArrowHead	&	251	&	3	&	36	&	175	&	MidPhalOutAgeGroup	&	80	&	3	&	400	&	154 \\
		Beef	&	470	&	5	&	30	&	30	&	MidPhalOutCorrect	&	80	&	2	&	600	&	291 \\
		BeetleFly	&	512	&	2	&	20	&	20	&	MiddlePhalanxTW	&	80	&	6	&	399	&	154 \\
		BirdChicken	&	512	&	2	&	20	&	20	&	MoteStrain	&	84	&	2	&	20	&	1252 \\
		Car	&	577	&	4	&	60	&	60	&	NonInvasiveThorax1	&	750	&	42	&	1800	&	1965 \\
		CBF	&	128	&	3	&	30	&	900	&	NonInvasiveThorax2	&	750	&	42	&	1800	&	1965 \\
		ChlorineConcentration	&	166	&	3	&	467	&	3840	&	OliveOil	&	570	&	4	&	30	&	30 \\
		CinCECGtorso	&	1639	&	4	&	40	&	1380	&	OSULeaf	&	427	&	6	&	200	&	242 \\
		Coffee	&	286	&	2	&	28	&	28	&	PhalOutCorrect	&	80	&	2	&	1800	&	858 \\
		Computers	&	720	&	2	&	250	&	250	&	Phoneme	&	1024	&	39	&	214	&	1896 \\
		CricketX	&	300	&	12	&	390	&	390	&	Plane	&	144	&	7	&	105	&	105 \\
		CricketY	&	300	&	12	&	390	&	390	&	ProxPhalOutAgeGroup	&	80	&	3	&	400	&	205 \\
		CricketZ	&	300	&	12	&	390	&	390	&	ProxPhalOutCorrect	&	80	&	2	&	600	&	291 \\
		DiatomSizeReduction	&	345	&	4	&	16	&	306	&	ProximalPhalanxTW	&	80	&	6	&	400	&	205 \\
		DisPhalOutAgeGroup	&	80	&	3	&	400	&	139	&	RefrigerationDevices	&	720	&	3	&	375	&	375 \\
		DisPhalOutCor	&	80	&	2	&	600	&	276	&	ScreenType	&	720	&	3	&	375	&	375 \\
		DislPhalTW	&	80	&	6	&	400	&	139	&	ShapeletSim	&	500	&	2	&	20	&	180 \\
		Earthquakes	&	512	&	2	&	322	&	139	&	ShapesAll	&	512	&	60	&	600	&	600 \\
		ECG200	&	96	&	2	&	100	&	100	&	SmallKitchApps	&	720	&	3	&	375	&	375 \\
		ECG5000	&	140	&	5	&	500	&	4500	&	SonyAIBORSurface1	&	70	&	2	&	20	&	601 \\
		ECGFiveDays	&	136	&	2	&	23	&	861	&	SonyAIBORSurface2	&	65	&	2	&	27	&	953 \\
		ElectricDevices	&	96	&	7	&	8926	&	7711	&	StarlightCurves	&	1024	&	3	&	1000	&	8236 \\
		FaceAll	&	131	&	14	&	560	&	1690	&	Strawberry	&	235	&	2	&	613	&	370 \\
		FaceFour	&	350	&	4	&	24	&	88	&	SwedishLeaf	&	128	&	15	&	500	&	625 \\
		FacesUCR	&	131	&	14	&	200	&	2050	&	Symbols	&	398	&	6	&	25	&	995 \\
		FiftyWords	&	270	&	50	&	450	&	455	&	SyntheticControl	&	60	&	6	&	300	&	300 \\
		Fish	&	463	&	7	&	175	&	175	&	ToeSegmentation1	&	277	&	2	&	40	&	228 \\
		FordA	&	500	&	2	&	3601	&	1320	&	ToeSegmentation2	&	343	&	2	&	36	&	130 \\
		FordB	&	500	&	2	&	3636	&	810	&	Trace	&	275	&	4	&	100	&	100 \\
		GunPoint	&	150	&	2	&	50	&	150	&	TwoLeadECG	&	82	&	2	&	23	&	1139 \\
		Ham	&	431	&	2	&	109	&	105	&	TwoPatterns	&	128	&	4	&	1000	&	4000 \\
		HandOutlines	&	2709	&	2	&	1000	&	370	&	UWaveAll	&	945	&	8	&	896	&	3582 \\
		Haptics	&	1092	&	5	&	155	&	308	&	UWaveX	&	315	&	8	&	896	&	3582 \\
		Herring	&	512	&	2	&	64	&	64	&	UWaveY	&	315	&	8	&	896	&	3582 \\
		InlineSkate	&	1882	&	7	&	100	&	550	&	UWaveZ	&	315	&	8	&	896	&	3582 \\
		InsectWingbeatSound	&	256	&	11	&	220	&	1980	&	Wafer	&	152	&	2	&	1000	&	6164 \\
		ItalyPowerDemand	&	24	&	2	&	67	&	1029	&	Wine	&	234	&	2	&	57	&	54 \\
		LargeKitchApps	&	720	&	3	&	375	&	375	&	WordSynonyms	&	270	&	25	&	267	&	638 \\
		Lightning2	&	637	&	2	&	60	&	61	&	Worms	&	900	&	5	&	181	&	77 \\
		Lightning7	&	319	&	7	&	70	&	73	&	WormsTwoClass	&	900	&	2	&	181	&	77 \\
		Mallat	&	1024	&	8	&	55	&	2345	&	Yoga	&	426	&	2	&	300	&	3000 \\
		Meat	&	448	&	3	&	60	&	60	&	 & & & &  \\
		\hline
	\end{tabular}
\end{table}

\section*{Appendix B: Reviews}

This paper was rejected after a second round of reviewing from the journal Machine Learning. We include the reviews for the interested reader.

Comments for the Author:

Reviewer 1: The authors have addressed my comments to my satisfaction. The changes in the paper in response to the other reviewers' comments are mostly in the direction of improvement. I found some of the authors' answers a bit rushed and awkwardly phrased.

Reviewer 2: The paper still concerns a large-scale evaluation of rotation forest, which I agree is underappreciated, but this paper does not offer any new evidence that was previously not known.
While the contribution is essentially two-fold: 1) contract version of the algorithm and 2) a large-scale empirical investigation, the former is very simplistic and the latter does not add any substantial new information about the rotation forest algorithm. The authors does not substantially manage to influence my earlier opinion about the paper. It is still very incremental and does not provide any additional insights over the papers identified in the literature.

Reviewer 4: Thank you very much for the opportunity to review your work. Your manuscript focuses on rotation forest, a tree-based ensemble classifier introduced by Rodriguez, Kuncheva \& Alonso (2006) that aims to increase ensemble member diversity through feature extraction. In its introducing paper, an extensive empirical validation using 33 data sets proved that rotation forest was superior over random forest, bagging and adaboost. I do agree with the authors rotation forest is an underestimated ensemble classifier that deserves more attention in literature. The manuscript investigates the performance of rotation forests in the setting where all features are continuous and provides an elaborate benchmark study where it is compared to a selection of state-of-the-art competitors. Furthermore, your study intends to shed light on the underlying mechanisms that explain rotation forest's superiority, in casu in comparison to random forest. Finally, inspired by the recognition that rotation forest is more complex in nature that random forest which is translated in longer training times, the authors present a method for estimating training times and adjusting the training procedure accordingly.

There are several points I appreciated whilst reading your work. Your experiments are very elaborate since they are based on a very large set of datasets and rotation forest is compared to many benchmark algorithms. Moreover, your experiments do explicitly recognize the importance of tuning hyperparameters using cross-validation. This is indeed especially relevant in a study that compares classifiers - you need to be vigilant that every algorithm is given a fair chance by either using default values for all parameters, or by utilizing its potential to its fullest by optimizing parameter values. And finally, you conduct an analyses into the underlying mechanisms of performance and suggest an algorithm variation that aims to reduce runtimes.
Unfortunately, I also found the study to be fundamentally flawed in a number of ways.

First, I have a major concern about the study's contribution and positioning. Although your contributions are not explicitly identified in the introduction, the main novelty of study you emphasize lies in the evaluation of rotation forest in the specific setting of datasets containing only continuous variables. Unfortunately the study currently fails to explain why this unusual, more narrow focus in comparison to other, domain-agnostic benchmark studies, is relevant. I miss a clear, literature-based justification early on in the manuscript that convinces your reader where such settings our found, why it is necessary to (re-)evaluate classifiers in this specific setting, or why one can expect classifiers to behave differently in a continuous-only versus a mixed-feature setting. In the absence of such a motivation, your study seems to offer little more than a replication of experiments found in previous work that has demonstrated the strong classification performance of rotation forest in comparison to other, well-established benchmarks over a variety of datasets.
Based on personal experience, having only continuous features in classification tasks is very exceptional. The observation that (i) only 39 of 121 UCI datasets meet this restriction, and (ii) the UCR data sets are fundamentally time-series classification datasets (for which cross-sectional techniques such as rotation or random forests are in principle not designed) seems to confirm this.
I see two potential solutions. One is that the manuscript (more specifically, the introduction) is re-written to provide a clear justification for this limitation of scope by referring to relevant literature that suggests classifiers behave differently in this setting, and by providing ample examples of domains where one could expect this type of data. The data sets selected for the empirical validation should be representative of these domains. Another solution would be to limit the scope of the study to an evaluation of rotation forest for time series classification.

Second, your manuscript is focusing heavily on (references to) specific coding and software implementations in Weka. You often refer to specific algorithm implementations, functions and function arguments in Java/Weka. While I find this highly unusual, this is unfortunate for a number of other reasons.
 
 First, it seems you are reducing the definition of the rotation forest algorithm to its specific implementation in Weka. While this code has been released by the original authors of the algorithm (Rodriguez, Kuncheva \& Alonso; 2006), it makes for a too restructure definition of the rotation forest algorithm. It is important to make a distinction between the essential design features of the classifier, and certain experimental settings that were adopted in their study and the Weka implementation. For example: you state that rotation forest "uses a C4.5 decision tree" for base learners. Actually, this is simply the base learner they used for their experiments - rotation forest algorithm is more broadly defined and can in principle be based on any base learner. Also in terms of feature extraction method, rotation forest is not exclusively bound to the usage of PCA. Follow-up work by Kuncheva \& Rodriguez (2007) has relaxed the definition as they experimented on the usage of alternative methods such as nonlinear discriminant analysis and random projections. Also the parameter that determines the class selection probability in random forest (hard-coded in Weka) is not fixed in the definition of random forest.

I think in empirical work it is always useful to explain in which software environment experiments were set up, and if so, which existing packages or libraries you depended on. Moreover, if new algorithms are proposed I believe it is in the interest of both the researcher (to gain more visibility) and the research community (to encourage usage and facilitate follow-up work) to release code. However, I think such a discussion should be limited to a short, dedicated section or paragraph in a discussion of experimental settings. Studies involving definitions of new algorithms and results of benchmark study with other methods should be written in a more generalized manner so that reliance upon a single software environment is avoided. I must admit that your paper often reads as a technical report on the usage of Weka for using rotation forest. In principle, your paper should not make any reference to software environments, Java classes or function arguments beyond a section on experimental settings.

Third, the paper does not sufficiently recognize and discuss earlier work on rotation forest and its alternative configurations in detail in Section 3. You seem to dismiss existing literature mainly since it is not restricted to datasets involving only numerical features. However, especially since the justification of the strictly numerical setting is missing (see point 1) I do not see how this literature is not relevant, nor could I detect a valid reason to not discuss these works in detail. Actually, earlier work already extensively investigated many design features for rotation forest: ensemble size (Liu \& Huang, 2008; Kuncheva \& Rodriguez, 2007), feature extraction algorithm (Liu \& Huang, 2008; De Bock \& Van den Poel, 2010; Kuncheva \& Rodriguez, 2007), the number of feature subsets (Kuncheva \& Rodriguez, 2007) and even its member classifiers . Many other, more recent papers, also optimize rotation forest in function of its algorithm components and hyperparameters. I suggest providing a very elaborate discussion of what has been investigated prior to your study, and in which settings. Adding a table with an overview of this body of literature seems the best way forward.

Finally, I think it is very interesting to investigate how a classifier can be simplified to reduce its runtime. This is basically your objective in the last section of the manuscript where you present the contract rotation forest variation. You present a method for estimating training times based on data set characteristics (number of instances and features) and rotation forest parameter settings. This is very interesting and I personally think this is where the real contribution of your study lies. Taking into account runtimes in algorithm comparisons is extremely challenging and can be easily criticized. For example, what is the role of the specific software and hardwire environment or coding of the algorithms' components? What about opportunity costs and trade-offs between tuning parameters, training larger, more complex classifiers, engineering features etc. However, there are a number of other problems in your current approach which severely impact its usefulness in my view.
 
 Throughout a number of analysis (Tables 5 and 6) you establish that the performance of rotation forest can be maintained whilst speeding up training times by introducing a random feature selection at the ensemble member level. But this analysis seems incomplete since there are additional dimensions that have a substantial impact on runtimes (i) other hyperparameters, such as the number of feature subsets (rotation forest) and the number of features per random subset (random forest). Moreover, in rotation forest, two additional sources that have a huge impact on runtimes are the chosen base classifier, and the feature extraction algorithm (which you rightfully establish on page 22). The observation that you do not mention these factors is related to your too narrow definition of rotation forests. This makes me question why an analyst would be interested in using your contract rotation forest by removing features or instances, potentially harming performance, if a similar runtime gain could be obtained by using alternative feature extraction method, such as random sparse projections (suggested in Kuncheva and Rodriguez, 2007), or using another base classifier.

Your contract rotation forest sets out to enhance runtimes in comparison to standard rotation forests, but also abandons your manuscript's focus on improving classifier performance. As you mention on page 29: "We stress that improved accuracy is not our goal: our intention is to improve usability". One can raise the question on the relevance of an algorithm that runs faster, but where this might result in a model that predicts less accurate (and where the drop in accuracy is unknown). Your algorithm does not include any mechanism for protecting accuracy or for managing the trade-off between runtimes and accuracy.
* Finally, the explanation of your algorithm is incomplete and confusing. This section requires substantial re-writing. Results are incomplete, since in contrast to the preceding sections you limit classification performance reporting to error rates. Additionally, to demonstrate the effectiveness of the reduction schemes in the algorithm, it would be interesting to provide results on (average) numbers of trees in the (reduced) ensembles, and average reductions of the numbers of instances and features.

Other comments

P.2 paragraphs 2-5: you motivate certain experimental decisions here (data set selection, parameter tuning, evaluation metrics and benchmark algorithm selection). I suggest moving this motivation to the section on experimental settings.
 
P.3: "the original description of the algorithm set the default number of trees to ten". Incorrent, the number of trees is a hyperparameter. In Rodriguez et al. (2006), experiments conducted limited the number of trees of all ensemble classifiers to ten. So this was an experimental decision.
 
 P.5: "RandomTree": this is a Weka-specific classifier designed to implement random forest. Actually, random forest depends on CART trees with random feature selection at the node level.

P. 5, 6 Your explanation of rotation forest contains a mistake. The correct parameter is the number of feature subsets, not the the number of features. Rodriguez et al. (2006) suggest a default value of 3. The same applies to their Weka code. Your Table 1 suggests to test 3 to 12 features per subset. That means that for datasets with more than 36 features, your values do not cover the recommended default split. The risk is that the results you reported for rotation forest are underestimations rather than the opposite, so this unfavorable choice is unlikely to cancel the findings in your experimental benchmark. However, it does raise the question of how your sensitivity analysis for this parameter (p.19) is relevant.
 
 P.7: "Many (UCI data sets) are very small, and have been formatted in an unusual way: categorical variables have been made real-valued in a way that may confound the classifiers". Actually, categorical values are usually dummy-coded (and thus not real-valued) to make them compatible with certain classifiers that are not tree-based. This is common practice and I don't see any reason why this would confound a classifier. Please provide more motivation or refer to a source to support this argument to discredit UCI data sets containing non-continuous features.

P.7: "Experiments are conducted by averaging over 30 stratified resamples of data". This is uncommon. A typical cross-validation consists of m replications of a n-fold cross-validation where common choices for m and n are (15,10), (10,10) or (5,2). Provide references to previous analogous approaches or justify, or re-run the experiments.

\end{document}